\documentclass[conference]{IEEEtran}

\usepackage{graphicx}
\usepackage{epstopdf}
\usepackage{subfigure} 
\usepackage{amsmath}
\usepackage{amsfonts}
\usepackage{amssymb}
\usepackage{amsthm}
\usepackage{mathrsfs}
\usepackage{algorithm}
\usepackage{algorithmic}
\usepackage{caption}
\usepackage{dsfont}

\theoremstyle{remark}

\newtheorem*{remark*}{Remark}

\DeclareMathOperator{\sign}{\mathrm{sign}}

\DeclareMathOperator*{\argmin}{\arg\min}
\DeclareMathOperator*{\argmax}{\arg\max}

\newcommand{\cI}{\mathcal{I}}

\newcommand{\cC}{\mathcal{C}}

\newcommand{\dOne}{\mathds{1}}

\begin{document}
%
\title{Task Selection for Bandit-Based Task Assignment in Heterogeneous Crowdsourcing}

\author{\IEEEauthorblockN{Hao Zhang}
\IEEEauthorblockA{Department of Computer Science\\
Tokyo Institute of Technology, Japan\\
zhang.h.ae@m.titech.ac.jp}
\and
\IEEEauthorblockN{Masashi Sugiyama}
\IEEEauthorblockA{Department of Complexity Science and Engineering\\
The University of Tokyo, Japan\\
sugi@k.u-tokyo.ac.jp}
}

\maketitle

\begin{abstract}
\emph{Task selection} (picking an appropriate labeling task) and \emph{worker selection} (assigning the labeling task to a suitable worker) are two major challenges in \emph{task assignment} for crowdsourcing. Recently, worker selection has been successfully addressed by the \emph{bandit-based task assignment} (BBTA) method, while task selection has not been thoroughly investigated yet. In this paper, we experimentally compare several task selection strategies borrowed from active learning literature, and show that the \emph{least confidence} strategy significantly improves the performance of task assignment in crowdsourcing.
\end{abstract}

\IEEEpeerreviewmaketitle

\section{Introduction}
\label{sec-intro}
Training labels are essentially important to machine learning tasks. Traditionally, training labels are collected from experts. However, this could be very expensive and time-consuming, especially when the unlabeled data is large-scale. The recent rise of \emph{crowdsourcing} \cite{howe08} has enabled us to efficiently combine human intelligence by asking a crowd of low-paid workers to complete a group of micro-tasks, such as \emph{image labeling} in computer vision \cite{welinder10a} and \emph{recognizing textual entailment} in natural language processing \cite{snow08}. Although crowdsourcing has provided us a cheaper and faster way for collecting labels, the collected labels are often (highly) noisy, since workers are usually non-experts. This is the trade-off between cost, time and quality. However, crowdsourcing services (e.g. \emph{Amazon Mechanical Turk}\footnote{https://www.mturk.com/mturk/}) are still successful, because there exists a phenomenon of the \emph{wisdom of crowds} \cite{liu13}: properly combining a group of untrained people can be as good as the experts in many application domains. Then how to properly combine the crowds is what to investigate in crowdsourcing research.

The primary challenge in crowdsourcing research is how to estimate the ground truth by using noisy labels from workers with various unknown reliability \cite{dawid79,raykar10,welinder10b,kajino12,liu12,zhou12}. Many existing methods are based on \emph{Expectation-Maximization} (EM) \cite{dempster77}, jointly learning workers' reliability and inferring true labels. We refer to these methods as ``static'' methods for crowdsourcing, as they usually run on the \emph{collected} labels but do not focus on \emph{how to collect} these labels. From the perspective of a requester, it is necessary to consider how to \emph{adaptively} collect labels from workers, for the purpose of intelligently using the total budget.

This motivates us to consider another important problem in crowdsourcing research, which is called \emph{task assignment} or task routing. In this paper, we focus on \emph{push} crowdsourcing marketplaces, where the system takes complete control over which labeling tasks are assigned to whom \cite{law11}. In contrast to ``static'' methods, we refer to task assignment methods for crowdsourcing as ``dynamic'' methods. Most of the existing task assignment methods run in an online mode, simultaneously learning workers' reliability and collecting labels \cite{donmez09,chen13,ertekin14}. They use different mechanisms to deal with the exploration (i.e. learning which workers are reliable) and exploitation (i.e. selecting the workers considered to be reliable) trade-off. For more details of these methods, we refer the reader to \cite[Section~4]{zhang15}.

In recent crowdsourcing marketplaces, the \emph{heterogeneity} of tasks is increasing. In such a scenario (we call it \emph{heterogeneous} crowdsourcing), a worker may be reliable at only a subset of tasks with a certain type. For example, when completing \emph{name entity recognition} tasks \cite{finin10,ritter11} in natural language processing, a worker may be good at recognizing names of sports teams, but not be familiar with cosmetics brands. Thus it is reasonable to model task-dependent reliability for workers in heterogeneous crowdsourcing. The task assignment methods mentioned above are designed for the \emph{homogeneous} setting, where they do not consider workers' reliability is task-dependent. Although they can also run in the heterogeneous setting, the performances could be poor (for experimental results, see \cite[Section~5]{zhang15}).

\emph{Bandit-based task assignment} (BBTA) \cite{zhang15} is a contextual bandit formulation for task assignment in heterogeneous crowdsourcing. In this formulation, a \emph{context} can be interpreted as the type or required skill of a task, and each arm of the bandit represents a worker. The feedback after pulling an arm depends on the current context. This corresponds to the task-dependent reliability of workers. BBTA focuses on the strategy of \emph{worker selection} in heterogeneous crowdsourcing. That is, given a task with a certain type, BBTA tries to select a suitable worker who tends to be good at this task.

As well as worker selection, \emph{task selection} (i.e. how to pick an appropriate task at each step) is also involved in the task assignment problem. A good strategy of task selection may help us efficiently use the budget when assigning the tasks to workers. For example, if we are already confident in the aggregated label of one certain task, we should not require more labels for this task, otherwise it would be a waste of budget. Although a common uncertainty criterion is used for picking tasks in BBTA, task selection is worth further investigating.

In this paper, we investigate the strategies of task selection for task assignment in heterogeneous crowdsourcing. In particular, we extend BBTA by using different strategies of picking tasks. The idea of these strategies is borrowed from \emph{query strategies} in \emph{active learning} \cite{settles09}. We embed several task selection strategies into BBTA, one of which is equivalent to the uncertainty criterion used in the original BBTA. We experimentally evaluate these strategies for BBTA, and demonstrate that the performance of BBTA can be further improved by adopting appropriate task selection strategies such as the \emph{least confidence} strategy.

\section{Bandit-Based Task Assignment}
\label{sec-bbta}
In this section, we review the method of \emph{bandit-based task assignment} (BBTA) \cite{zhang15}. The notation introduced in this section is used throughout the whole paper.

\subsection{Problem Setting and Notation}
We assume there are $N$ unlabeled tasks, $K$ workers, and $S$ contexts. The task is indexed by $i$, where $i\in[N]=\{1,2,\ldots,N\}$. The worker is indexed by $j$, where $j\in[K]=\{1,2,\ldots,K\}$. Each task is characterized by a context $s$, where $s\in[S]=\{1,2,\ldots,S\}$. For simplicity, we consider binary labels, i.e., the label space $\cC=\{-1,+1\}$. Each time given a task, we ask one worker from a pool of $K$ workers for a (possibly noisy) label, consuming one unit of the total budget $T$. The goal is to find suitable task-worker assignment to collect as many reliable labels as possible within the limited budget $T$. Finally, we estimate the true labels by aggregating the collected labels.

Let $y_{i,j}$ be the individual label of task $i$ (with context $s$) given by worker $j$. If the label is missing (i.e. we did not collect the label of task $i$ from worker $j$), we set $y_{i,j}=0$. We denote the weight of worker $j$ for context $s$ by $w^s_{j}$, corresponding to the task-dependent reliability of worker $j$ for context $s$. Note that $w^s_{j}$ is positive and dynamically learned in the method. Then an estimate of the true label is calculated by using the weighted voting mechanism as 
\begin{align}
\label{eq:weighted}
\widehat{y}_i=\sign\left(\frac{\sum_{j=1}^K w^s_{j}y_{i,j}}{\sum_{j'=1}^K w^s_{j'}}\right).
\end{align}
To avoid complex notation, we do not use the subscript for $s$, which implicitly represents the context of the current task $i$.

BBTA consists of the \emph{pure exploration} phase and the \emph{adaptive assignment} phase. The details are explained below.

\subsection{Pure Exploration Phase}
Pure exploration performs in a batch mode. The purpose is to partially know which workers are reliable at which labeling tasks. To this end, we pick $N'$ tasks for each of $S$ distinct contexts ($SN' \ll N$) and let \emph{all} $K$ workers label them. We denote the index set of $N'$ tasks with context $s$ by $\cI^s_1$ in this phase.

Since we have no prior knowledge of workers' reliability at this moment, we treat them equally and give all of them the same weight when aggregating their labels (equivalent to majority voting):
\begin{align*}
	\widehat{y}_i=\sign\left(\frac{1}{K}\sum_{j=1}^K y_{i,j}\right).
\end{align*}
In the standard crowdsourcing scenario, all of the true labels are usually unknown. As in many other crowdsourcing methods, we have the prior belief that most workers perform reasonably well. To evaluate an individual label $y_{i,j}$, using the weighted vote $\widehat{y}_i$ is a common solution \cite{donmez09,ertekin14}. We denote the cumulative loss by $L^s_{j,0}$ and initialize it for the next phase as
\begin{align*}
L^s_{j,0}=\sum_{i\in\cI^s_1}\dOne_{y_{i,j}\neq \widehat{y}_i},\text{ for }j\in[K]\text{ and }s\in[S],
\end{align*}
where $\dOne_{\pi}$ denotes the indicator function that outputs 1 if condition $\pi$ holds and 0 otherwise. This means that when a worker gives an individual label $y_{i,j}$ inconsistent or consistent with the weighted vote $\widehat{y}_i$, this worker suffers a loss 1 or 0. It is easy to see that cumulative losses correspond to workers' reliability. They are used for calculating workers' weights in the next phase. The budget for the next phase is $T_2=T-T_1$, where $T_1=SKN'<T$ is the budget consumed in this phase.

\subsection{Adaptive Assignment Phase}
In the adaptive assignment phase, task-worker assignment is determined for the remaining $N-SN'$ tasks in an online mode within the remaining budget $T_2$. At each step $t$ of this phase, to determine a task-worker pair, we need to further consider which task to pick and which worker to select for this task.

According to the weighted voting mechanism (Eqn.~\ref{eq:weighted}), we define the \emph{confidence score} for task $i$ as
\begin{align}
\label{eq:score}
	\overline{y}_i=\left|\frac{\sum_{j=1}^K w^s_{j}y_{i,j}}{\sum_{j'=1}^K w^s_{j'}}\right|.
\end{align}
Here, we use the different notation of the confidence score from that in the original paper of BBTA \cite{zhang15}, for the convenience of the unified notation in this paper. It is easy to see $\overline{y}_i\in[0,1]$. If the confidence score of a task is lower than those of others, collecting more labels for this task is a reasonable solution. Thus we pick task $i_t$ with the lowest confidence score as the next one to label:
\begin{align*}
i_t=\argmin_{i\in\cI_2}\overline{y}_i,
\end{align*}
where $\cI_2$ is the index set of current available tasks in this phase.

Given the picked task $i_t$ with context $s$, selecting a worker reliable at this task is always favored. On the other hand, workers' reliability is what we are dynamically learning in the method. We then use a contextual bandit \cite{bubeck12} formulation to handle this trade-off between exploration (i.e. learning which worker is reliable) and exploitation (i.e. selecting the worker considered to be reliable) in worker selection. Specifically, we calculate the weights of workers as follows:
\begin{align*}
w^s_{j,t^s}=\exp(-\eta^s_{t^s}L^s_{j,t^s-1}),\text{ for }j\in[K],
\end{align*}
where $t^s$ is the appearance count of context $s$ and $\eta^s_{t^s}$ is the learning rate related to $t^s$. This calculation of weights by using cumulative losses is due to the \emph{exponential weighting scheme} \cite{cesa-bianchi06,arora12}, which is a standard tool for sequential decision making. Following the exponential weighting scheme, we then select a worker $j_t$ from the discrete probability distribution on workers with each probability $p_{j,t}$ proportional to the weight $w^s_{j,t^s}$.

Then we ask worker $j_t$ for an individual label $y_{i_t,j_t}$ and calculate the weighted vote $\widehat{y}_{i_t}$ by using Eqn.~\ref{eq:weighted}. With the weighted vote $\widehat{y}_{i_t}$, we obtain the loss of the selected worker $j_t$: $l_{j_t,t}=\dOne_{\widehat{y}_{i_t}\neq y_{i_t,j_t}}$. Note that we can only observe the loss of the selected worker $j_t$, and for other workers, we decided to give an unbiased estimate of loss: $\widetilde{l}_{j,t}=\frac{l_{j,t}}{p_{j,t}}\dOne_{j=j_t}.$ 

Finally, we update the cumulative losses: $L^s_{j,t^s}=L^s_{j,t^s-1}+\widetilde{l}_{j,t}$ for $j\in[K]$, and the confidence scores of tasks with the same context as the current one.

The above assignment step is repeated $T_2$ times until the budget is used up.

For more details (including the pseudo code, regret analysis, comparison experiments), we refer the interested reader to the original paper of BBTA \cite{zhang15}.

\section{Task Selection Strategies for BBTA}
\label{sec-query}
In this section, we investigate how the performance of BBTA changes when we adopt different strategies of picking tasks.

\subsection{Task Selection Strategies}
The idea of task selection strategies in this paper is borrowed from query strategies, which are the common frameworks for measuring the \emph{informativeness} of instances in active learning \cite{settles09}. For example, at each step of an active learning algorithm, the most \emph{uncertain} instance is sampled for training, where the \emph{uncertainty} is calculated according to some query strategy.

Usually, query strategies in active learning are designed for \emph{probabilistic} models. In this paper, we introduce these strategies into our task assignment problem, where the weighted voting mechanism is \emph{non-probabilistic}. Although we do not have \emph{probabilities} of labels as in probabilistic models, we instead define a confidence score for each task, to describe how confident we are in the aggregated label of this task.

For convenience, we rewrite the weighted voting mechanism as
\begin{align*}
\widehat{y}_i=\argmax_{c\in\{-1,+1\}}\frac{\sum_{j=1}^{K}w^s_{j}\cdot\dOne_{y_{i,j}=c}}{\sum_{j'=1}^{K}w^s_{j'}},
\end{align*}
which is equivalent to Eqn.~\ref{eq:weighted}. We further define the \emph{positive score} and the \emph{negative score} respectively as
\begin{align*}
\overline{y}_i^+=\frac{\sum_{j=1}^{K}w^s_{j}\cdot\dOne_{y_{i,j}=+1}}{\sum_{j'=1}^{K}w^s_{j'}}\text{ and }\overline{y}_i^-=\frac{\sum_{j=1}^{K}w^s_{j}\cdot\dOne_{y_{i,j}=-1}}{\sum_{j'=1}^{K}w^s_{j'}}.
\end{align*}

It is easy to see that $\overline{y}_i^+,\overline{y}_i^-\in[0,1]$. For task $i$, the positive/negative score will be $1$ if and only if we have collected labels from \emph{all} workers and \emph{all} of them are consistent. On the other hand, when we have no positive/negative label for task $i$, the positive/negative score will be $0$.

At each step of the adaptive assignment phase, we pick a task according to the \emph{confidence score} $\overline{y}_i$, which is calculated by using the positive score and the negative score based on the specific task selection strategy we adopt. The details of task selection strategies are presented below.

\subsection{Least Confidence}
The strategy called \emph{least confidence} (LC) corresponds to probably the most commonly used framework of \emph{uncertainty sampling} in active learning \cite{lewis94}. 

Specifically, at step $t$ of the adaptive assignment phase, we pick the task with the index:
\begin{align*}
i_t=\argmin_{i\in\cI_2}\overline{y}_i,\text{ where }\overline{y}_i=\max(\overline{y}_i^+,\overline{y}_i^-).
\end{align*}

This is a straightforward way to pick a task. Here, the confidence score $\overline{y}_i$ is simply set as the larger one between the positive score $\overline{y}_i^+$ and the negative score $\overline{y}_i^-$. That is, it is the score of the more probable label ($-1$ or $+1$). Then we pick the task $i_t$ with the lowest confidence score, indicating that among all tasks we are the least confident in the aggregated label of task $i_t$, and thus we require one more label for this task at step $t$. 

\subsection{Margin Sampling}
The strategy called \emph{Margin sampling} (MS) is originally used for multi-class uncertainty sampling in active learning \cite{scheffer01}. Instead of only considering the information about the most probable label (as in LC), MS also involves the information about the remaining label(s). 

We introduce this strategy into our task assignment problem as
\begin{align*}
i_t=\argmin_{i\in\cI_2}\overline{y}_i,\text{ where }\overline{y}_i=\left|\overline{y}_i^+-\overline{y}_i^-\right|.
\end{align*}

It is easy to see that this strategy is equivalent to the uncertainty criterion (Eqn.~\ref{eq:score}) used in the original BBTA method. Basically, the confidence score $\overline{y}_i$ here is the absolute value of the difference between the sum of (normalized) weights for positive labels and that for negative labels. The uncertainty in MS can be interpreted as ambiguity. The intuition is that tasks with smaller margins are more ambiguous, and ambiguous tasks usually require more labels.

\begin{figure*}[t]
	\centering
	\subfigure[$N=351,K=30,S=3$]{
		\includegraphics[width=0.31\textwidth]{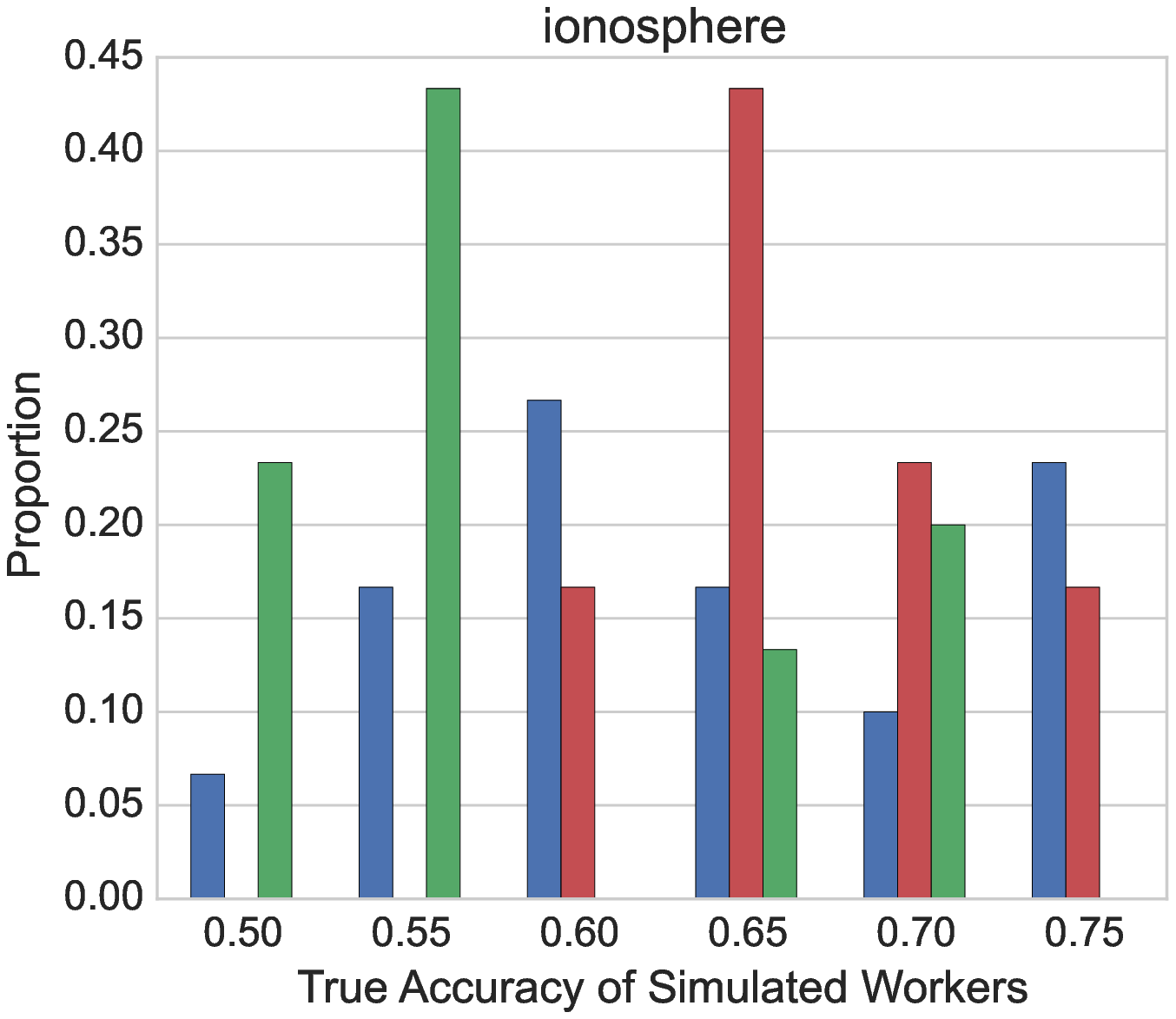}
		\label{fig:ionosphere-workers}
	}
	\subfigure[$N=569,K=40,S=4$]{
		\includegraphics[width=0.31\textwidth]{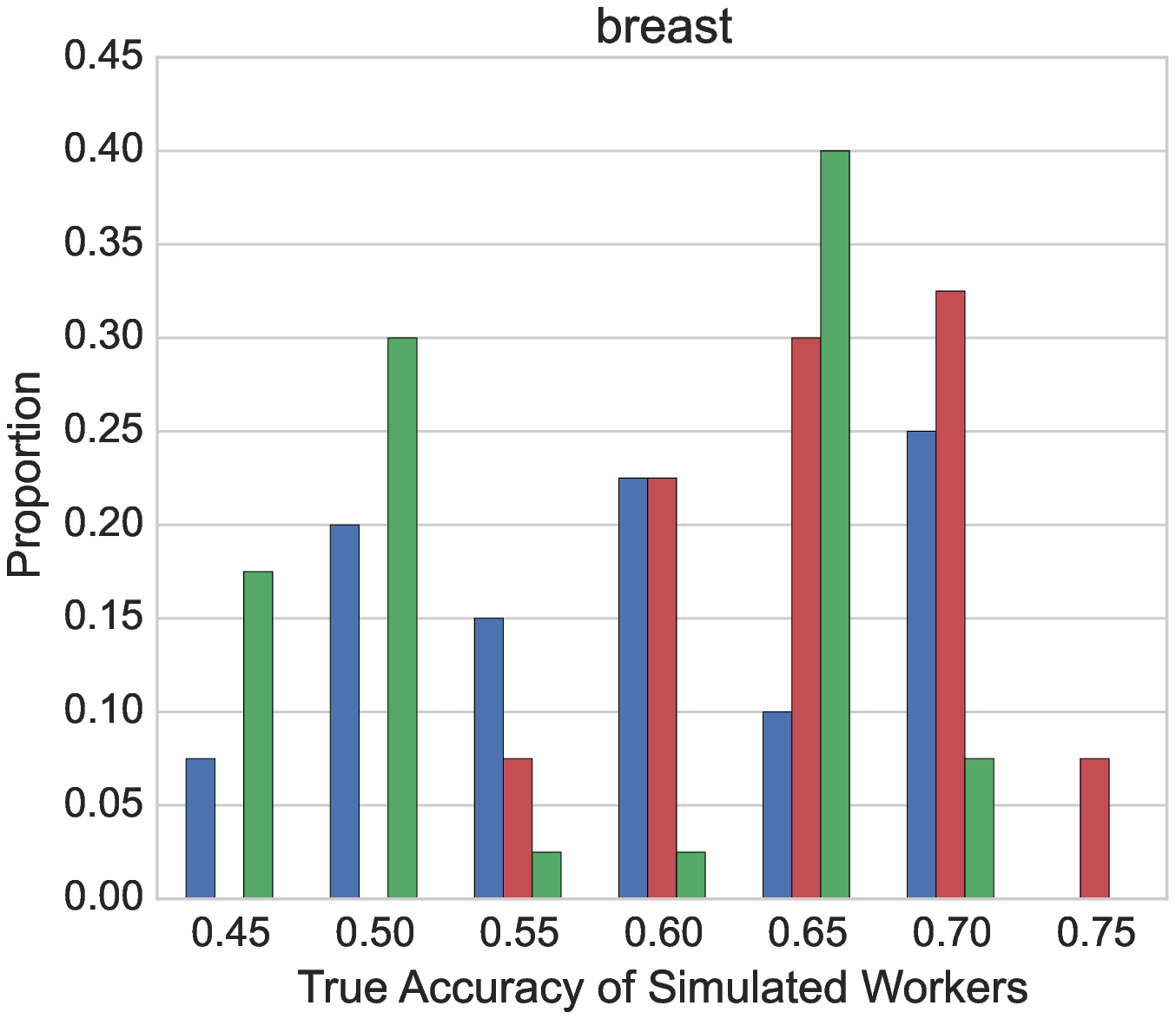}
		\label{fig:breast-workers}
	}
	\subfigure[$N=768,K=50,S=5$]{
		\includegraphics[width=0.31\textwidth]{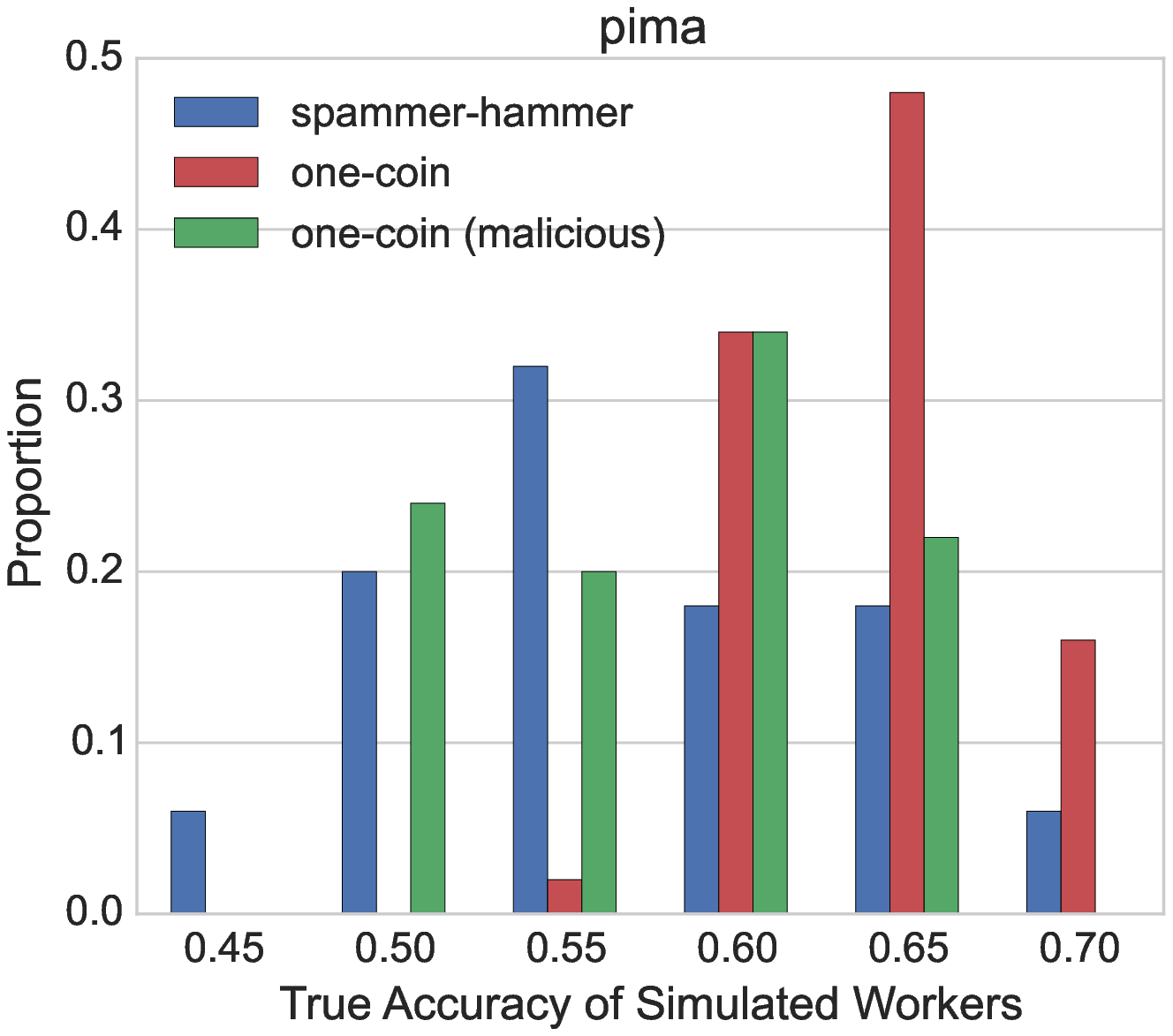}
		\label{fig:pima-workers}
	}
	\caption{Distribution of true accuracy of simulated workers for three benchmark datasets with three worker models.}
	\label{fig:bench-workers}
\end{figure*}

\subsection{Information Density}
It is argued in \cite{settles08} that the informativeness of instances should not only be measured by the uncertainty, but also by the \emph{representativeness} of the underlying distribution. This is the motivation of the strategy called \emph{Information Density} (ID) \cite{settles08}. 

Similarly, in our heterogeneous crowdsourcing setting, we can also consider the proportions of different task types as well as the confidence scores when picking tasks. We adopt the idea of ID and develop the task selection strategy as
\begin{align*}
i_t=\argmax_{i\in\cI_2}(1-\overline{y}_i)\left(\frac{N^s}{N}\right)^\beta,
\end{align*}
where $\overline{y}_i$ is calculated as in LC or MS, $s$ is the type (i.e. context) of task $i$, and $N^s$ is the total number of tasks with type $s$. The parameter $\beta$ controls the relative importance of the density term (i.e. the proportion of tasks with context $s$). When $\beta$ approaches 0, ID strategy degrades to LC or MS (depending on how we calculate $\overline{y}_i$). As $\beta$ gets larger, the density term becomes more important.

\section{Experiments}
\label{sec-exp}
In this section, we experimentally evaluate different task selection strategies for BBTA. Since the margin sampling strategy is equivalent to the uncertainty criterion in the original BBTA, for the purpose of fair comparison, we use the same experimental setup as that in the original paper of BBTA \cite{zhang15}.

We first conduct experiments on benchmark data with simulated workers, and then use real data for further comparison. All of the experimental results are averaged over 30 runs.

\subsection{Benchmark Data}

\begin{figure*}[t]
	\centering
	\subfigure[$N=351,K=30,S=3$]{
		\includegraphics[width=0.31\textwidth]{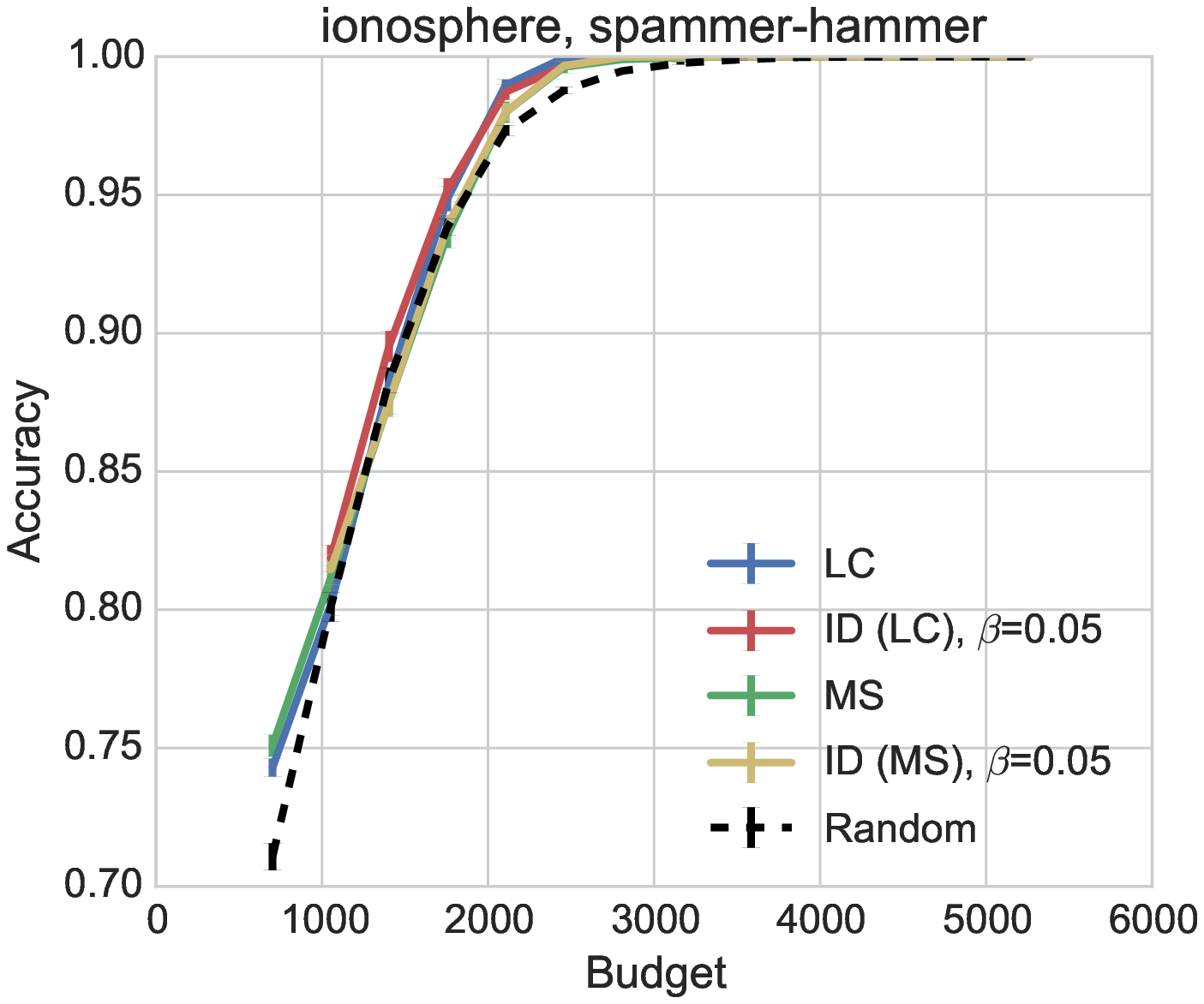}
		\label{fig:ionosphere_spammer}
	}
	\subfigure[$N=569,K=40,S=4$]{
		\includegraphics[width=0.31\textwidth]{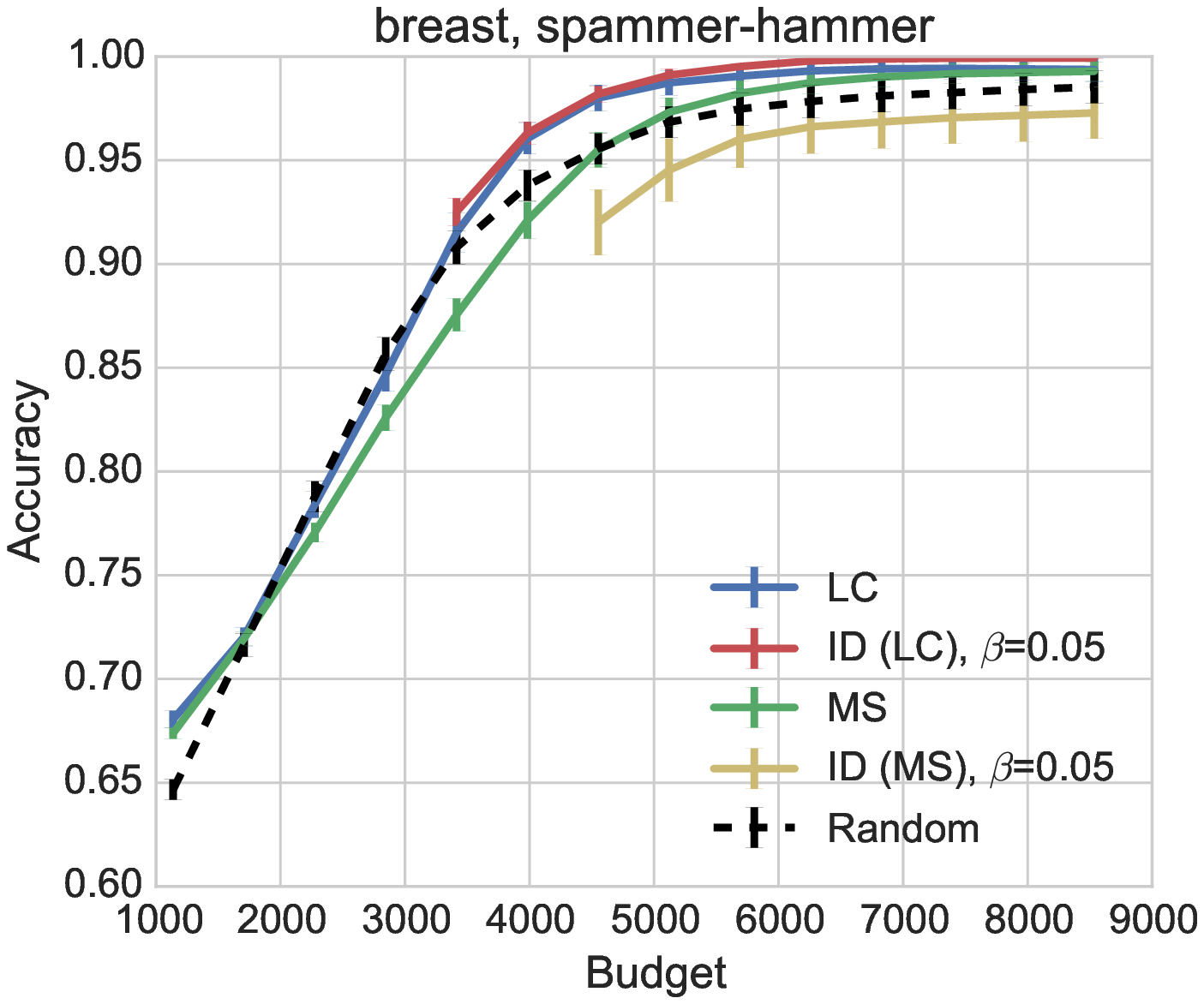}
		\label{fig:breast_spammer}
	}
	\subfigure[$N=768,K=50,S=5$]{
		\includegraphics[width=0.31\textwidth]{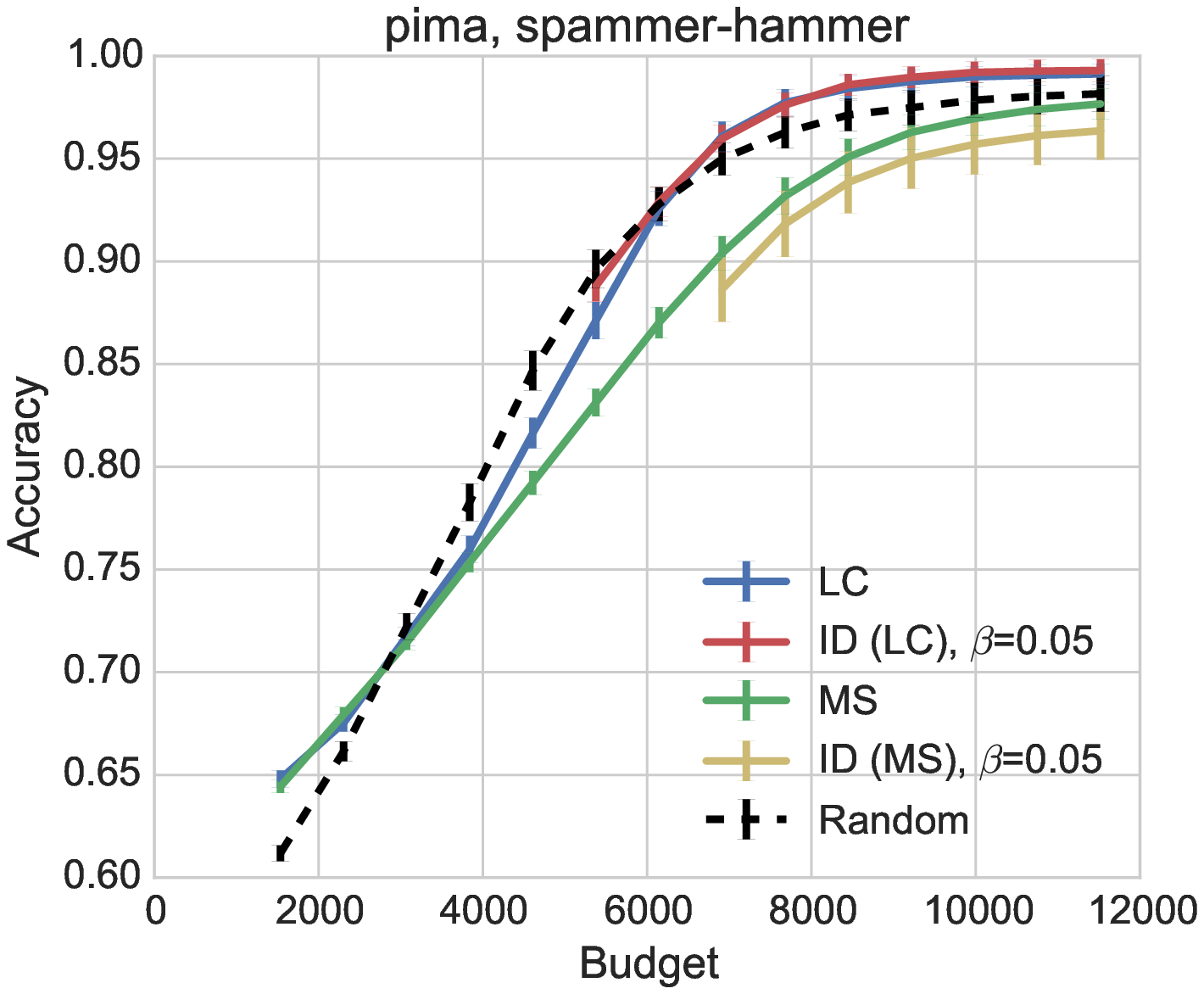}
		\label{fig:pima_spammer}
	}
	\subfigure[$N=351,K=30,S=3$]{
		\includegraphics[width=0.31\textwidth]{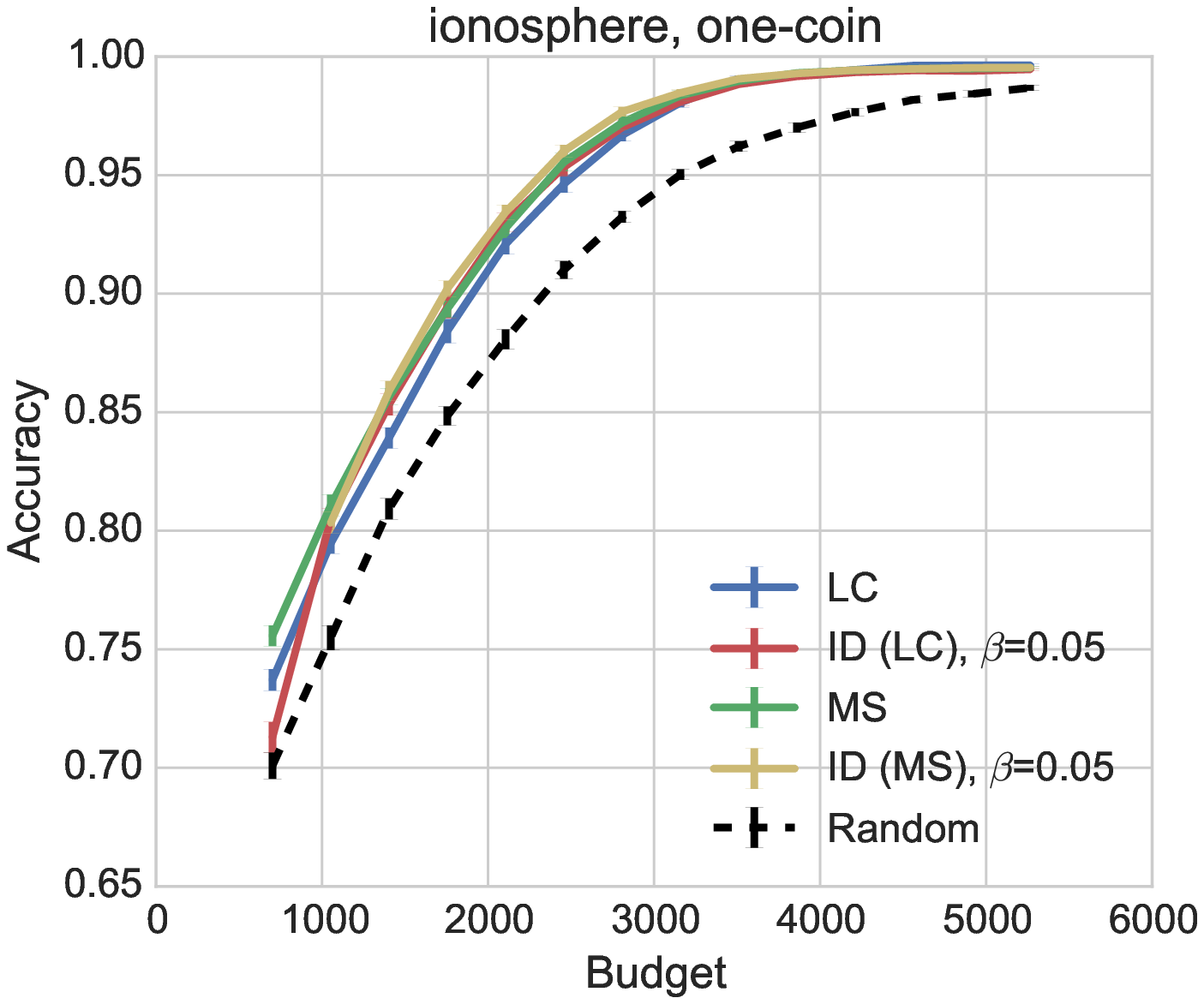}
		\label{fig:ionosphere_onecoin}
	}
	\subfigure[$N=569,K=40,S=4$]{
		\includegraphics[width=0.31\textwidth]{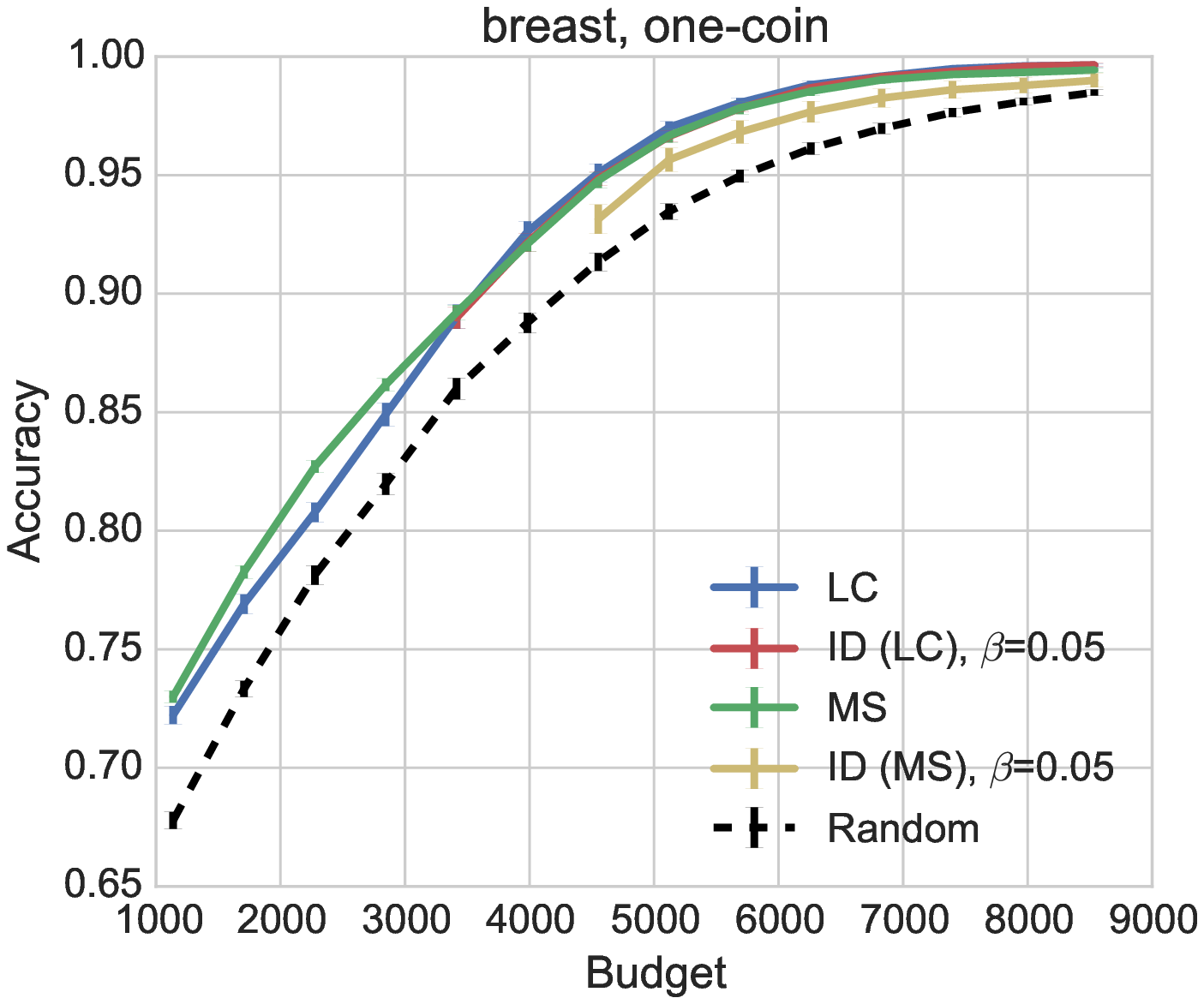}
		\label{fig:breast_onecoin}
	}
	\subfigure[$N=768,K=50,S=5$]{
		\includegraphics[width=0.31\textwidth]{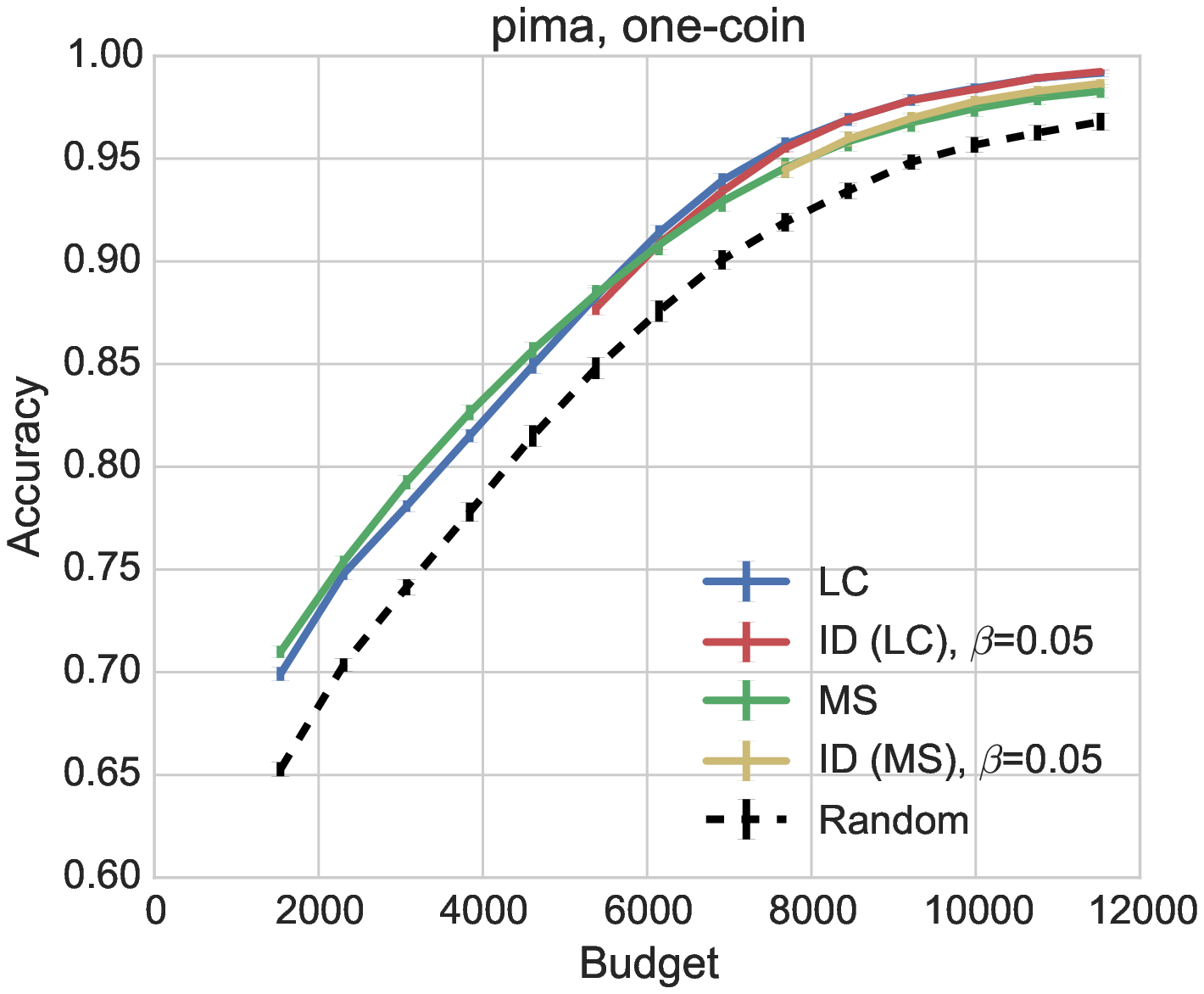}
		\label{fig:pima_onecoin}
	}
	\subfigure[$N=351,K=30,S=3$]{
		\includegraphics[width=0.31\textwidth]{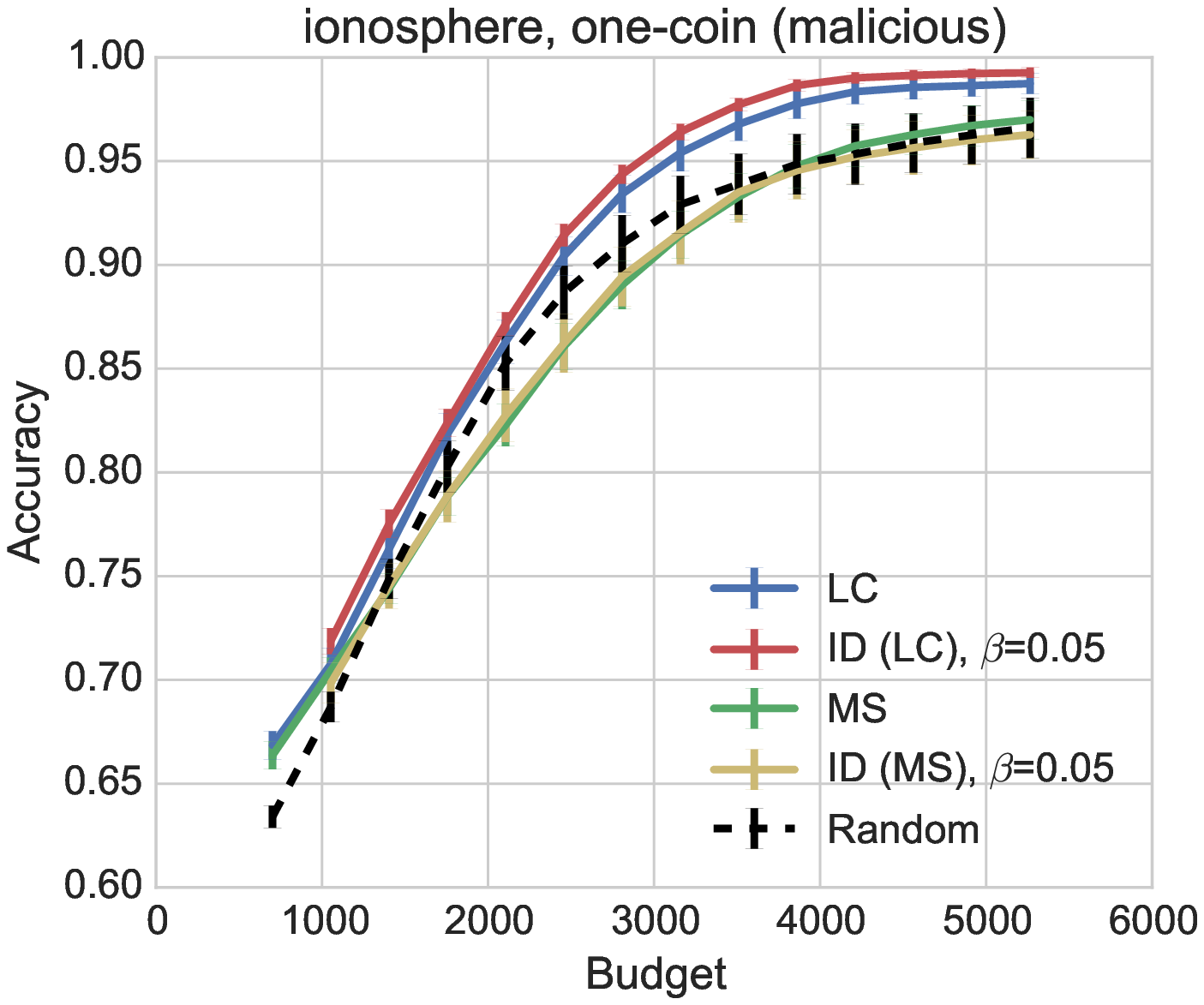}
		\label{fig:ionosphere_onecoin_m}
	}
	\subfigure[$N=569,K=40,S=4$]{
		\includegraphics[width=0.31\textwidth]{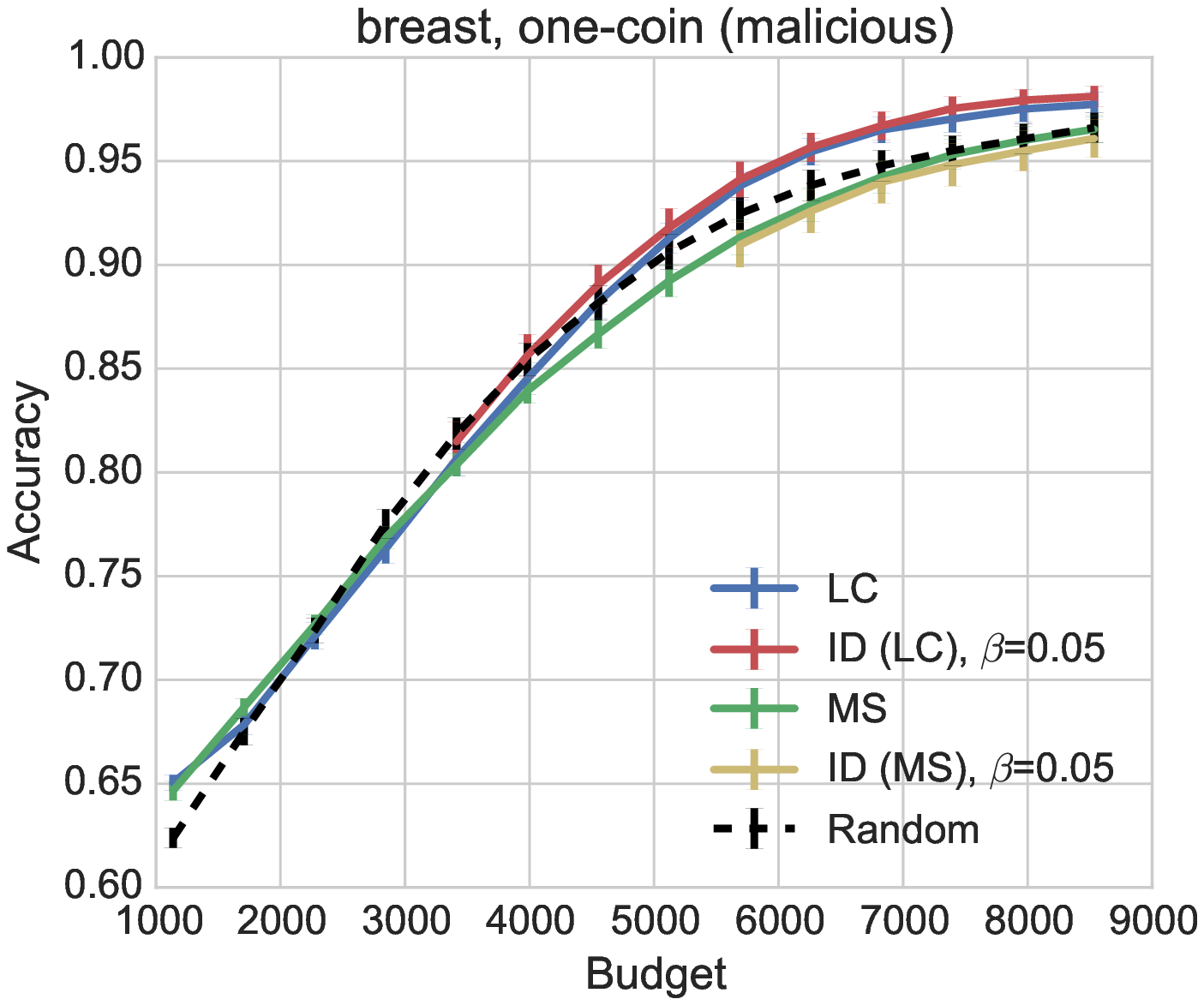}
		\label{fig:breast_onecoin_m}
	}
	\subfigure[$N=768,K=50,S=5$]{
		\includegraphics[width=0.31\textwidth]{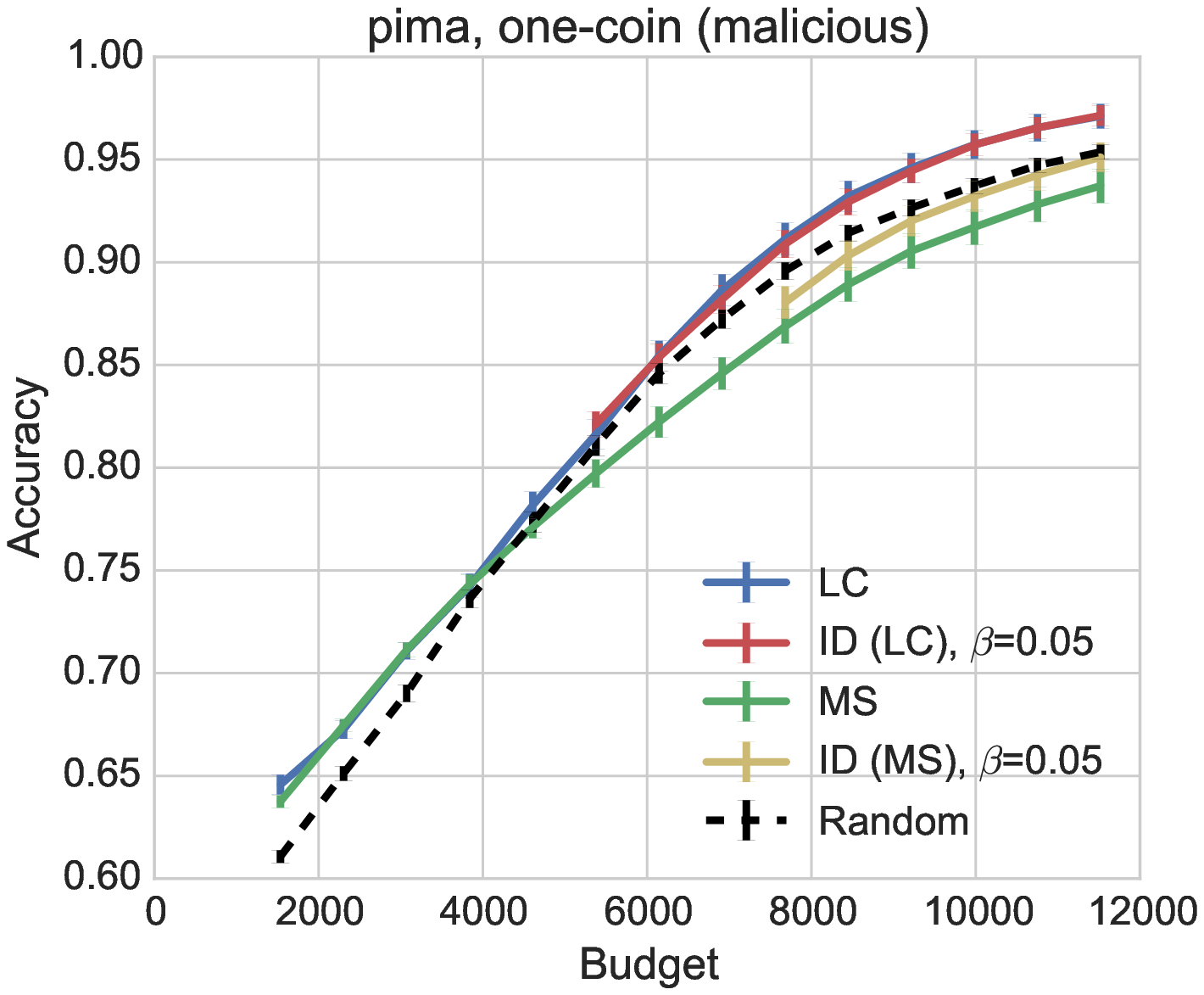}
		\label{fig:pima_onecoin_m}
	}
	\caption{Results of different task selection strategies on three benchmark datasets with three worker models.}
	\label{fig:bench-query-comparison}
\end{figure*}

\begin{figure*}[t]
	\centering
	\subfigure[$N=351,K=30,S=3$]{
		\includegraphics[width=0.31\textwidth]{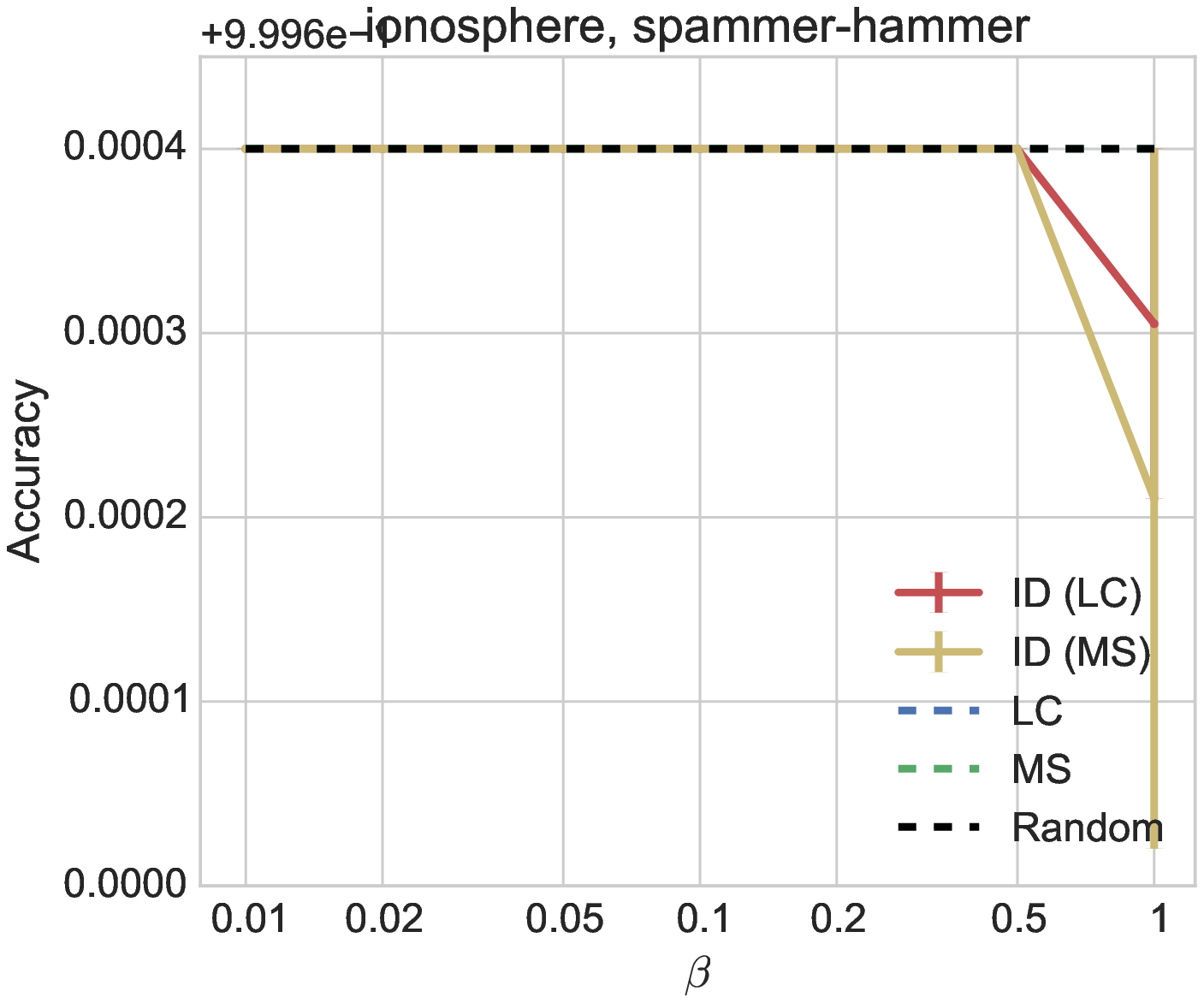}
		\label{fig:ionosphere_spammer-beta}
	}
	\subfigure[$N=569,K=40,S=4$]{
		\includegraphics[width=0.31\textwidth]{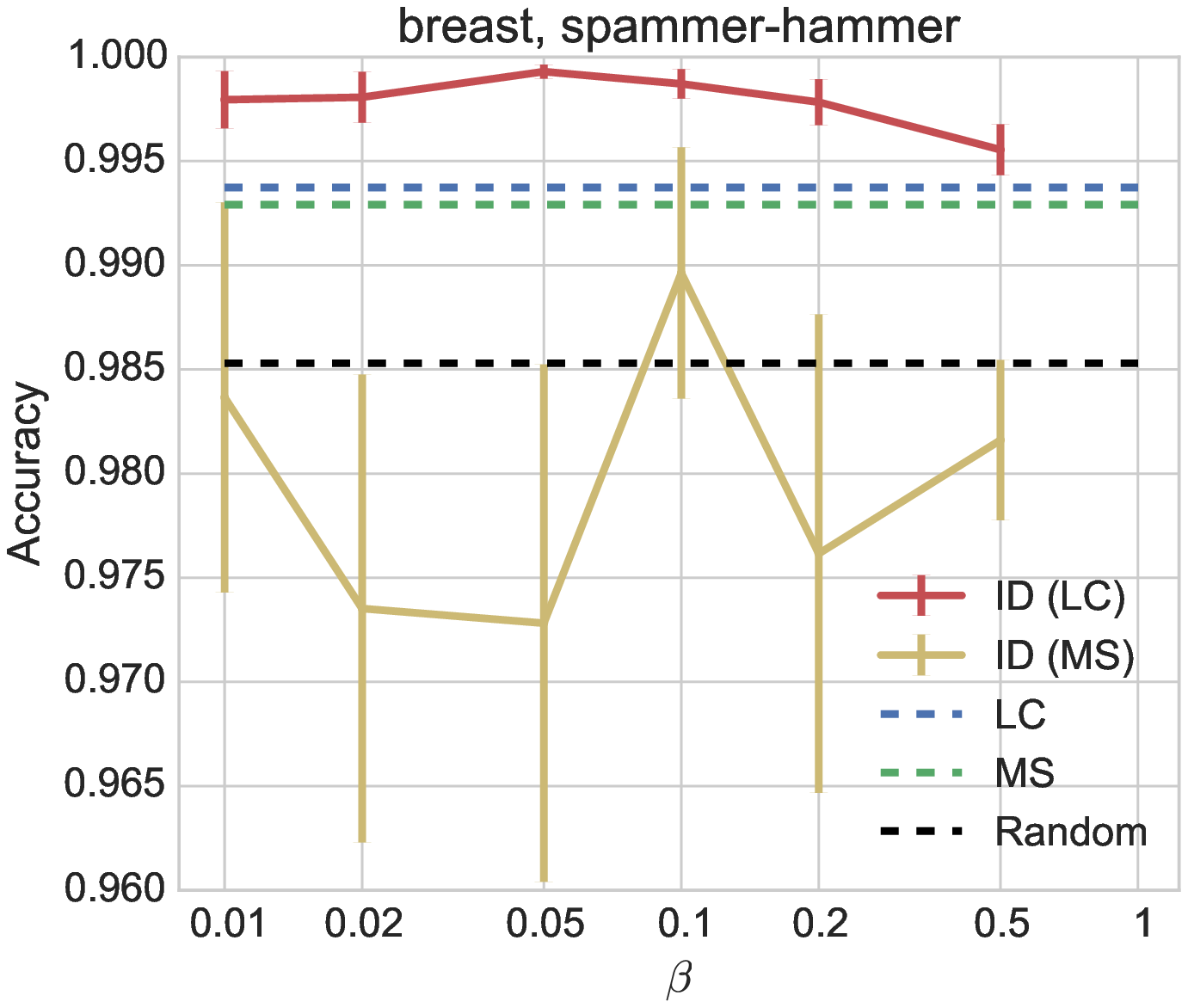}
		\label{fig:breast_spammer-beta}
	}
	\subfigure[$N=768,K=50,S=5$]{
		\includegraphics[width=0.31\textwidth]{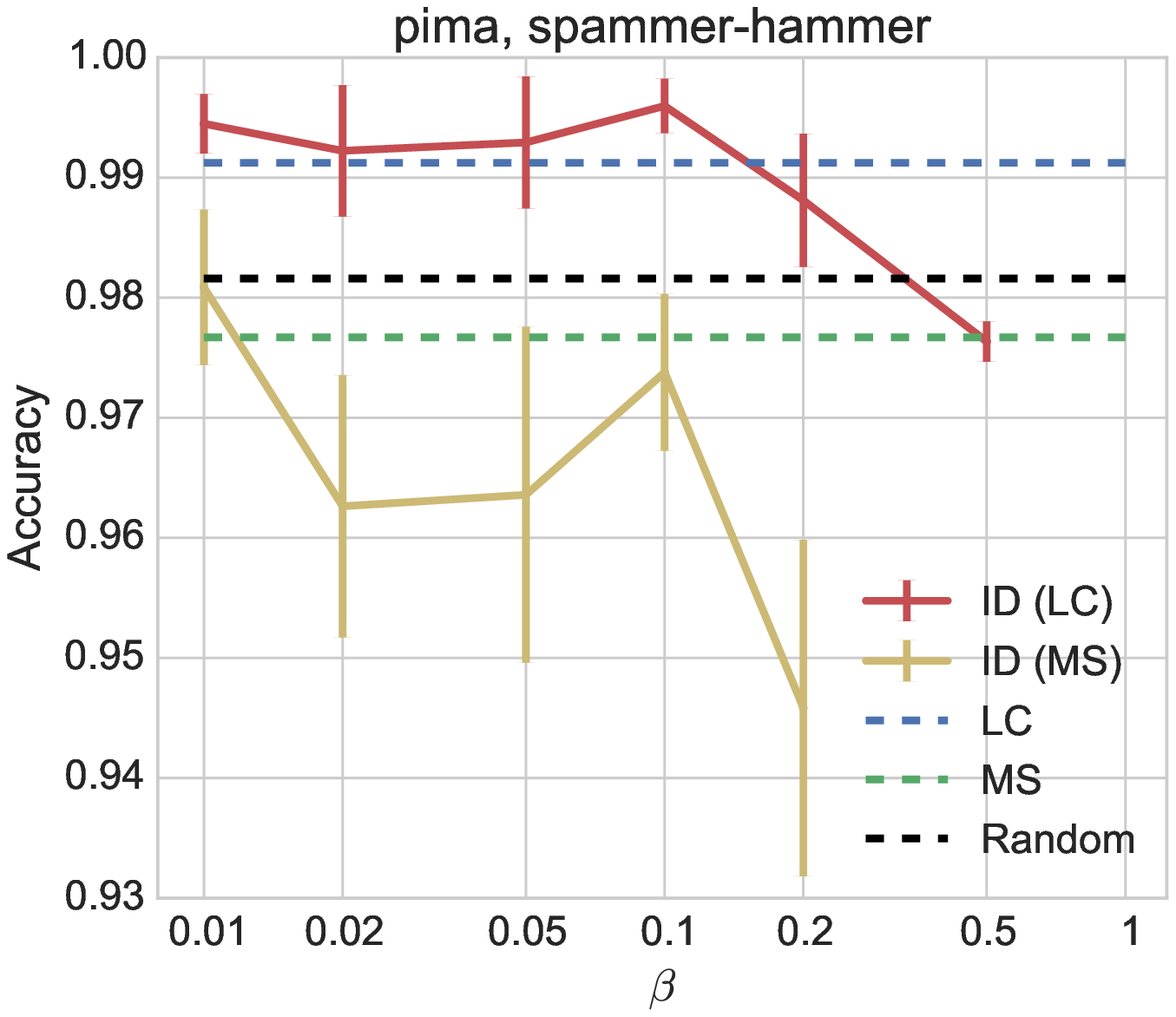}
		\label{fig:pima_spammer-beta}
	}
	\subfigure[$N=351,K=30,S=3$]{
		\includegraphics[width=0.31\textwidth]{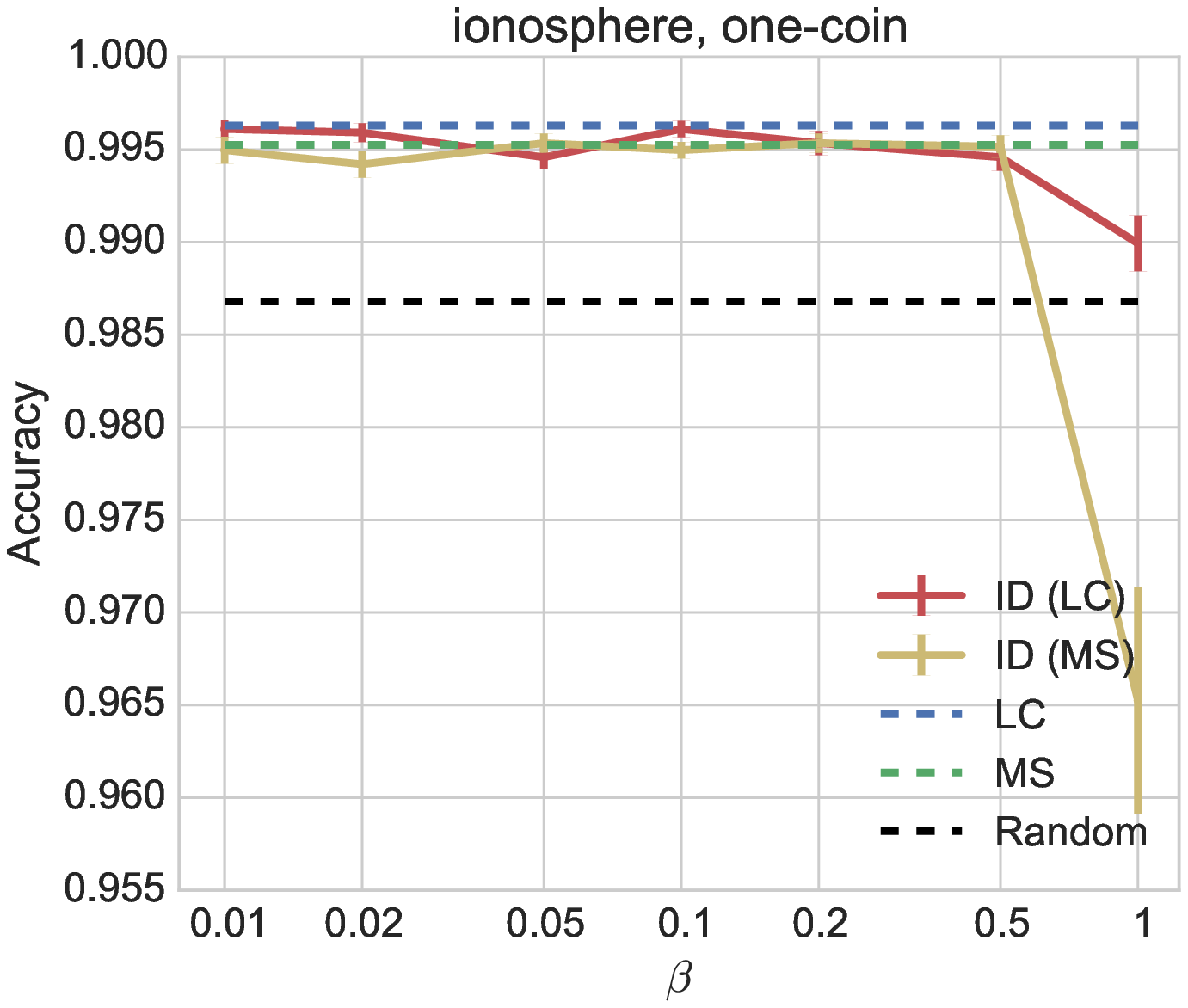}
		\label{fig:ionosphere_onecoin-beta}
	}
	\subfigure[$N=569,K=40,S=4$]{
		\includegraphics[width=0.31\textwidth]{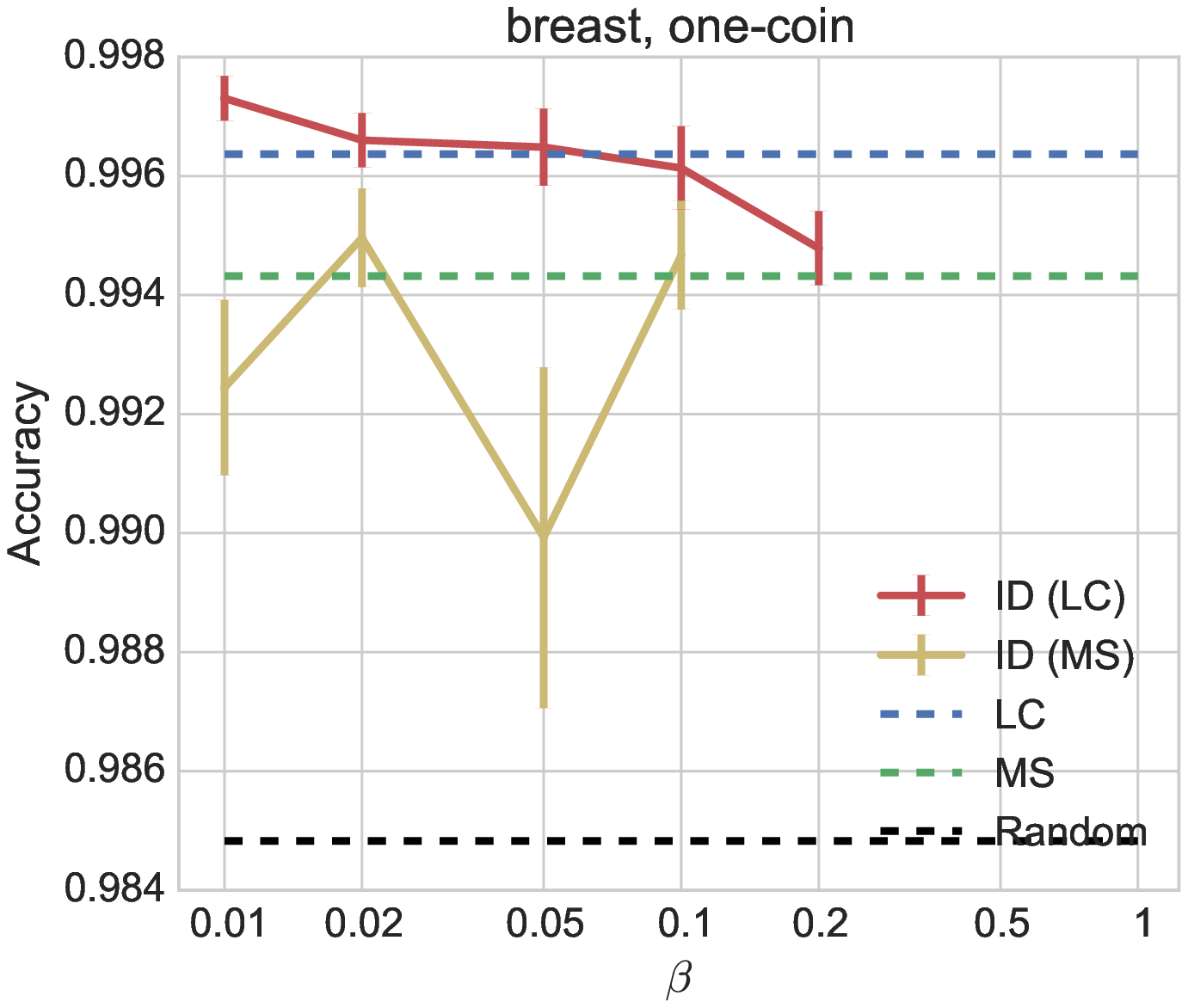}
		\label{fig:breast_onecoin-beta}
	}
	\subfigure[$N=768,K=50,S=5$]{
		\includegraphics[width=0.31\textwidth]{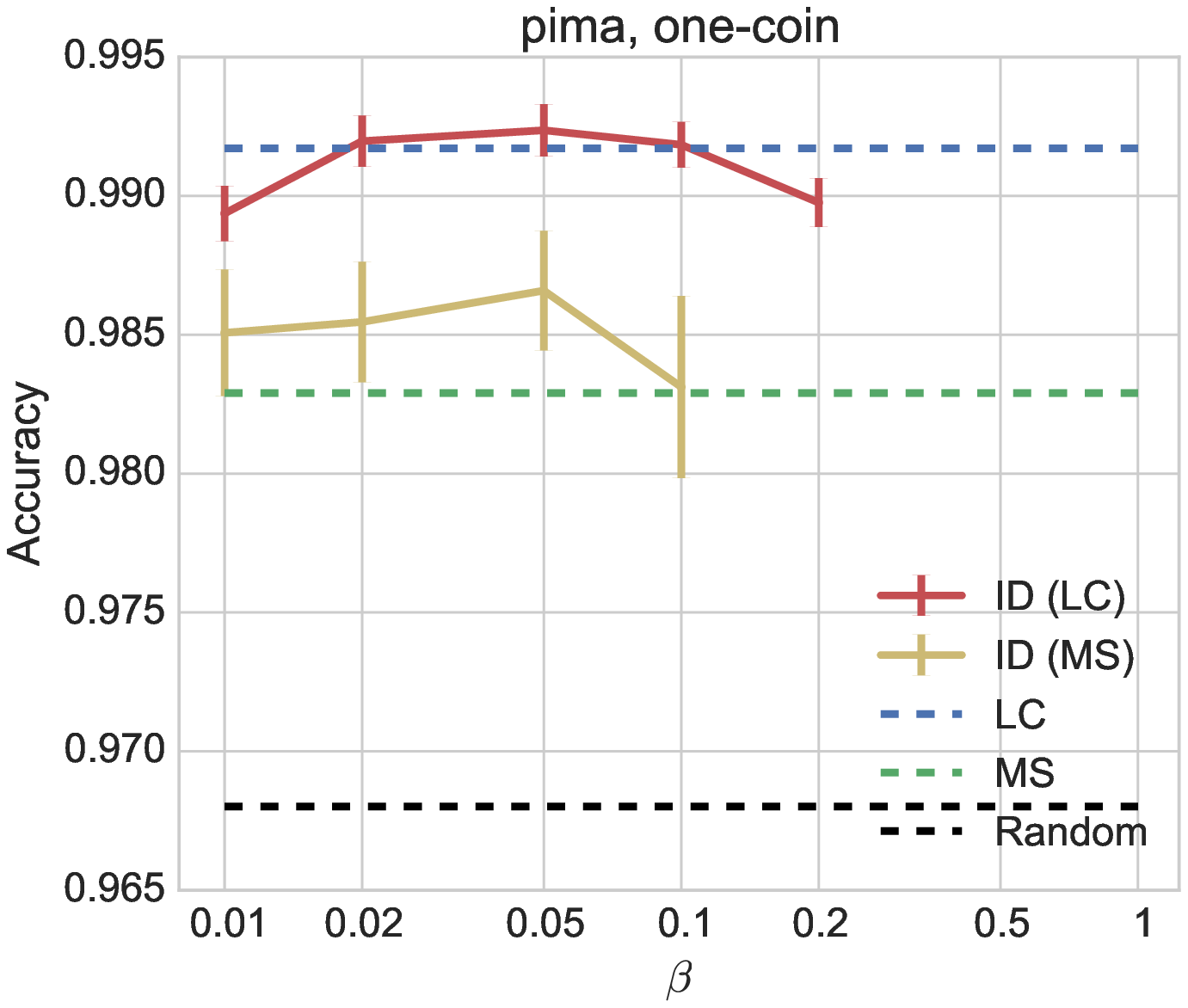}
		\label{fig:pima_onecoin-beta}
	}
	\subfigure[$N=351,K=30,S=3$]{
		\includegraphics[width=0.31\textwidth]{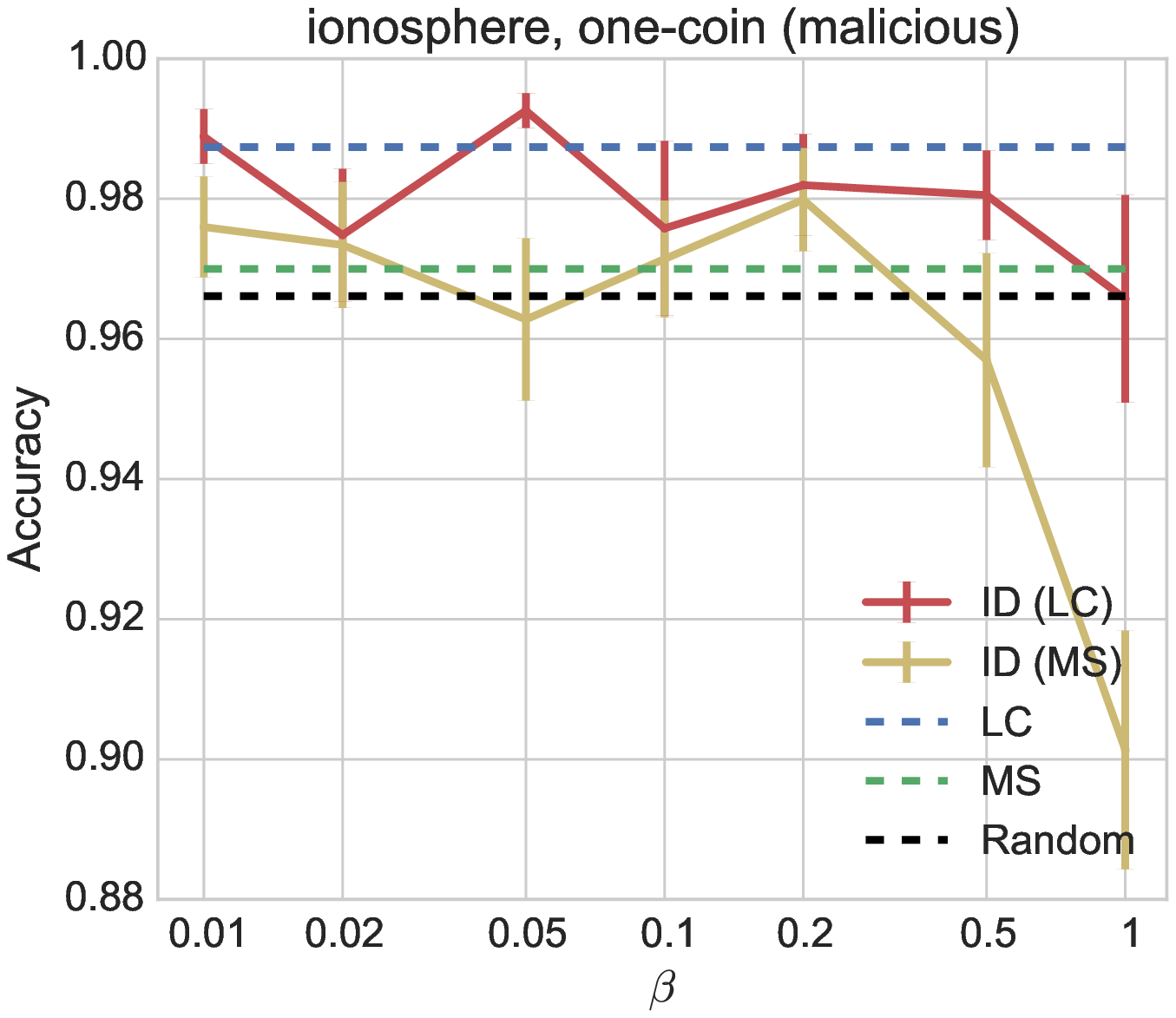}
		\label{fig:ionosphere_onecoin_m-beta}
	}
	\subfigure[$N=569,K=40,S=4$]{
		\includegraphics[width=0.31\textwidth]{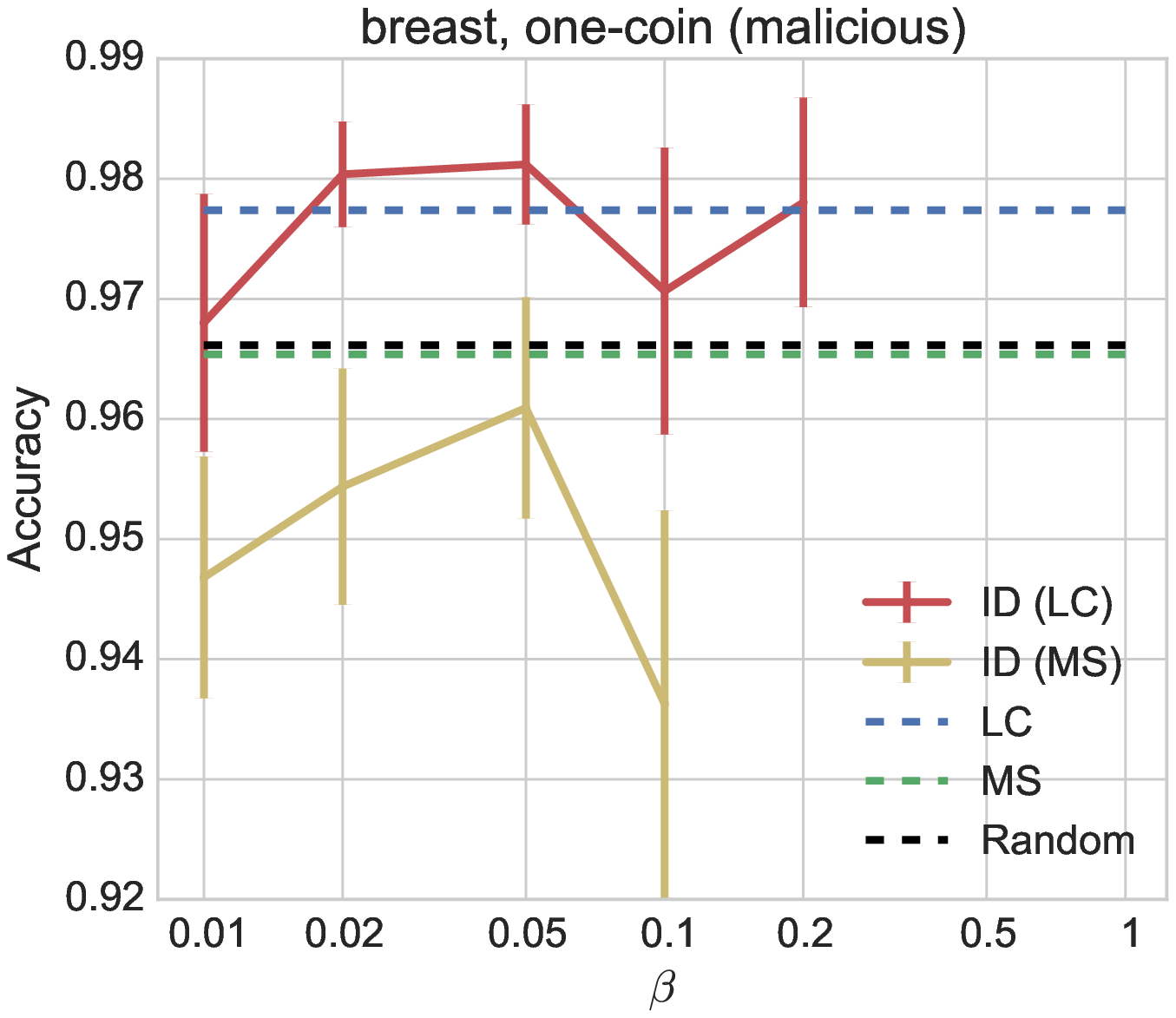}
		\label{fig:breast_onecoin_m-beta}
	}
	\subfigure[$N=768,K=50,S=5$]{
		\includegraphics[width=0.31\textwidth]{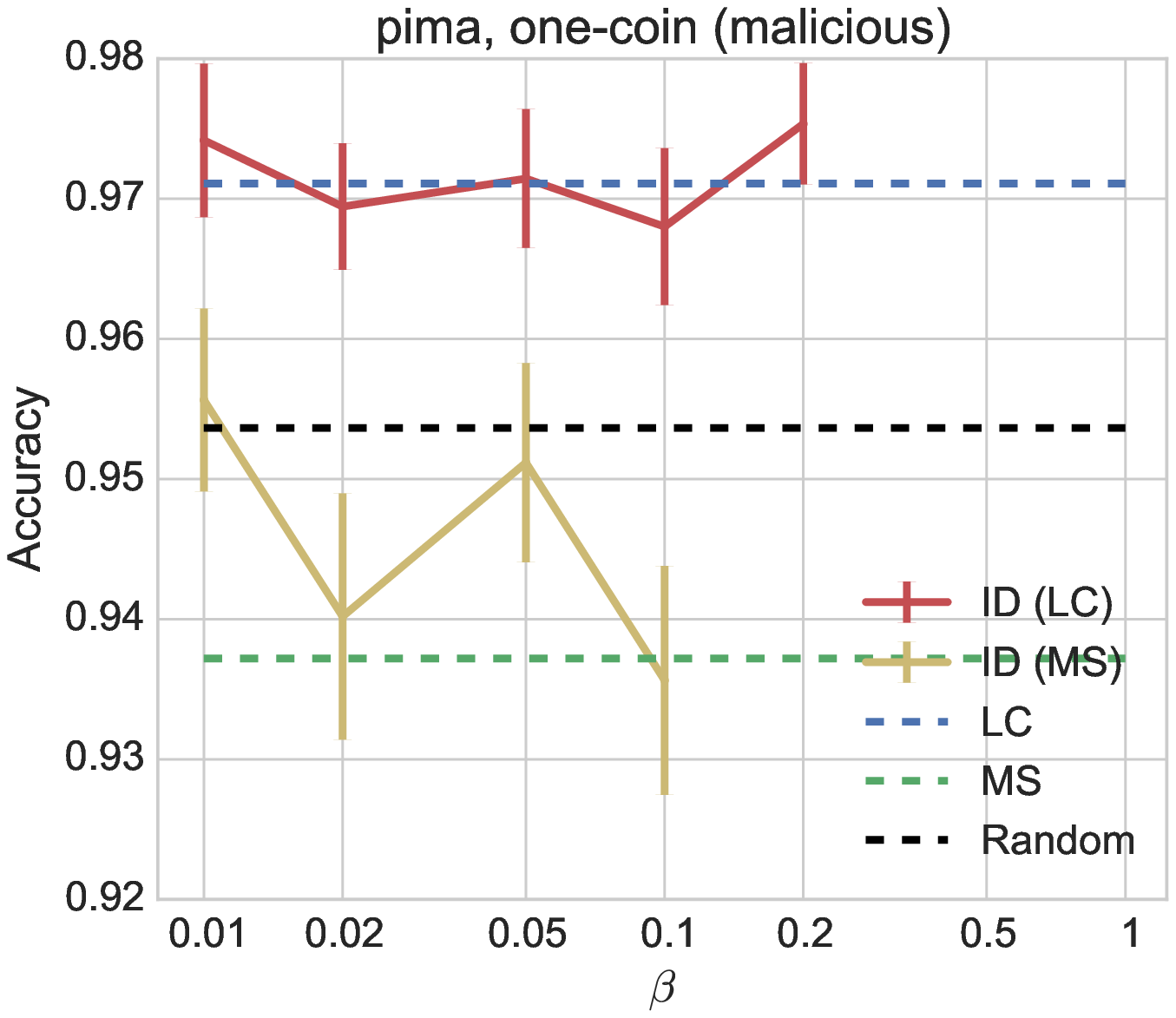}
		\label{fig:pima_onecoin_m-beta}
	}
	\caption{Results of varying $\beta$ in ID strategy on three benchmark datasets with three worker models.}
	\label{fig:bench-query-beta}
\end{figure*}

We perform experiments on three popular UCI benchmark datasets\footnote{http://archive.ics.uci.edu/ml/}: ionosphere ($N=351$), breast ($N=569$), and pima ($N=768$). We consider instances in these datasets as labeling tasks in crowdsourcing. True labels of all tasks in these datasets are available. To simulate various heterogeneous cases in the real world, we first use \emph{k-means} to cluster these three datasets into $S=3,4,5$ subsets respectively (corresponding to different contexts). Since there are no crowd workers in these datasets, we then simulate workers ($K=30,40,50$, respectively) by using the following worker models in heterogeneous setting:\\
\textbf{Spammer-Hammer Model:} A hammer gives true labels, while a spammer gives random labels \cite{karger11}. We introduce this model into the heterogeneous setting: each worker is a hammer on one subset of tasks (with the same context) but a spammer on others.\\
\textbf{One-Coin Model:} Each worker gives true labels with a given probability (i.e. accuracy). This model is widely used in many existing crowdsourcing literatures (e.g. \cite{raykar10,chen13}) for simulating workers. We use this model in heterogeneous setting: each worker gives true labels with higher accuracy (we set it to 0.9) on one subset of tasks, but with lower accuracy (we set it to 0.6) on others.\\
\textbf{One-Coin Model (Malicious):} This model is based on the previous one, except that we add more malicious labels: each worker is good at one subset of tasks (accuracy: 0.9), malicious or bad at another one (accuracy: 0.3), and normal at the rest (accuracy: 0.6).

With the generated labels from simulated workers, we can calculate the true accuracy for each worker by checking the consistency with the true labels. Figure \ref{fig:bench-workers} illustrates the proportions of simulated workers with the true accuracy falling in the associated interval (e.g., 0.65 represents that the true accuracy is between 60\% and 65\%). It is shown that the spammer-hammer model and the one-coin model (malicious) create more unreliable environments than those by the one-coin model.

We compare the least confidence (LC), margin sampling (MS) and information density (ID) strategies for BBTA in terms of accuracy. Accuracy is calculated as the proportion of correct estimates for true labels. We set $N'=1$ for all strategies. Recall that MS is equivalent to the uncertainty criterion used in the original BBTA method. For ID, we have ID (LC) and ID (MS), indicating that we use LC and MS respectively to calculate confidence scores. We set the parameter $\beta$ to 0.05 in both ID (LC) and ID (MS), which is a reasonable choice as shown later. We also implement a random strategy as a baseline, where we randomly pick a task at each step without using any uncertainty criterion. Accuracy of all strategies is compared at different levels of budget. We set the maximum amount of budget at $T=15N$.

Figure \ref{fig:bench-query-comparison} shows the averages and standard errors of accuracy as functions of budgets for all strategies in nine cases (i.e. three datasets with three worker models). Generally, LC and ID (LC) can be considered as the best two on these benchmark data, while ID (LC) is slightly better than LC in some cases. Surprisingly, MS and ID (MS) perform even worse than the random strategy in many cases. This implies that the uncertainty criterion used in the original BBTA may not be an appropriate task selection strategy.

We further check the effects of varying the parameter $\beta$ in ID. Figure~\ref{fig:bench-query-beta} shows the results. The accuracy is calculated when we consume all the budget, corresponding to the rightmost points on the curves in Figure~\ref{fig:bench-query-comparison}. For some larger $\beta$, we do not have the associated accuracy. This is because there exist some tasks that have never been assigned to any worker when $\beta$ is larger. We consider these cases as \emph{assignment failure} and do not calculate the accuracy for them. We also plot LC, MS and the random strategy for reference (dash curves without error bars). Generally, ID (LC) is better than ID (MS). Roughly speaking, the good choice of $\beta$ appears to be in the range between 0.01 and 0.1. 

\subsection{Real Data}
\begin{figure*}[t]
	\centering
	\subfigure[$N=800,K=164,S=1$]{
		\includegraphics[width=0.4\textwidth]{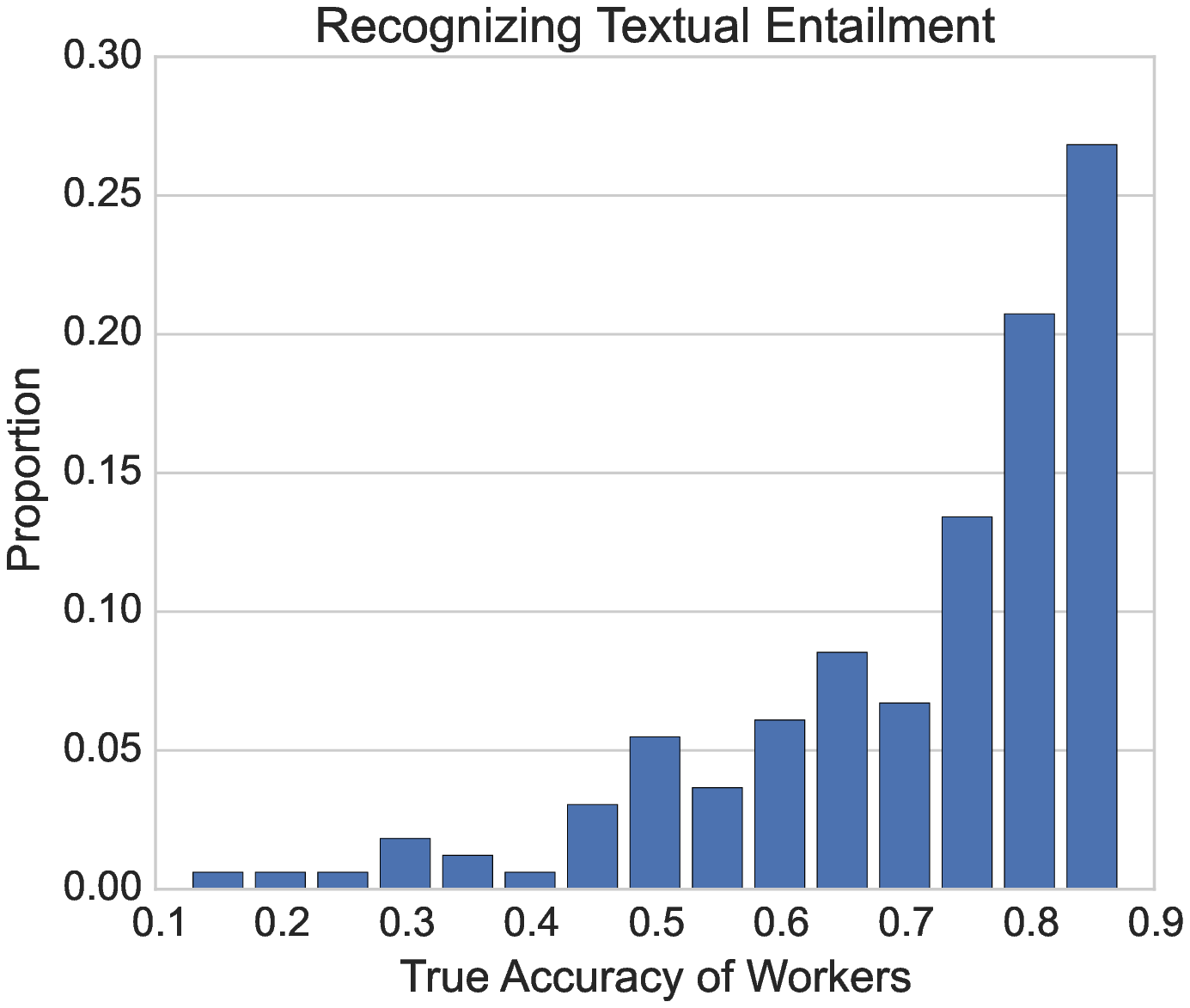}
		\label{fig:rte-workers}
	}
	\subfigure[$N=204,K=42,S=2$]{
		\includegraphics[width=0.4\textwidth]{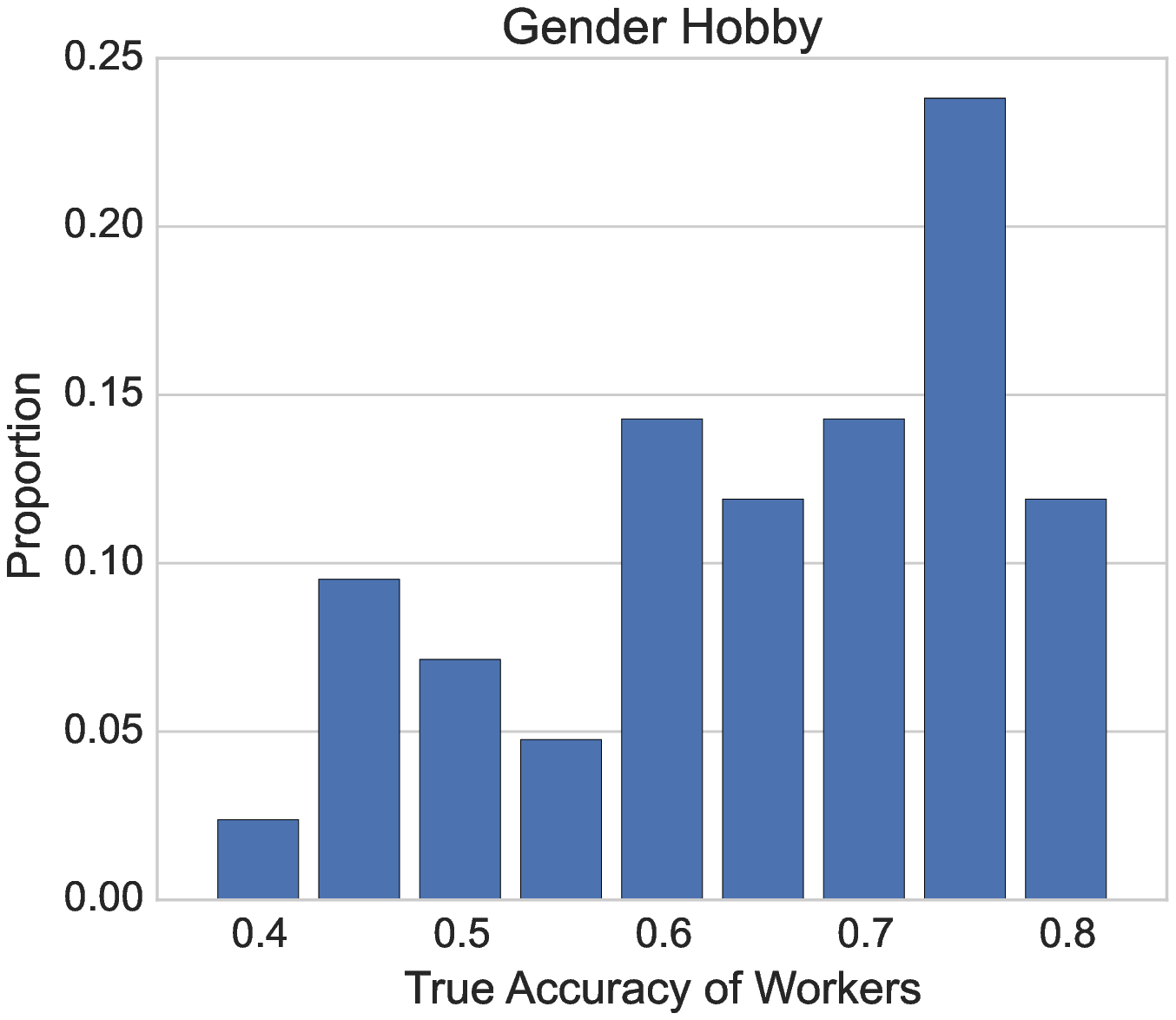}
		\label{fig:gh-workers}
	}
	\caption{Distribution of true accuracy of workers for two real-world datasets.}
	\label{fig:real-workers}
\end{figure*}

\begin{figure*}[t]
	\centering
	\subfigure[$N=800,K=164,S=1$]{
		\includegraphics[width=0.4\textwidth]{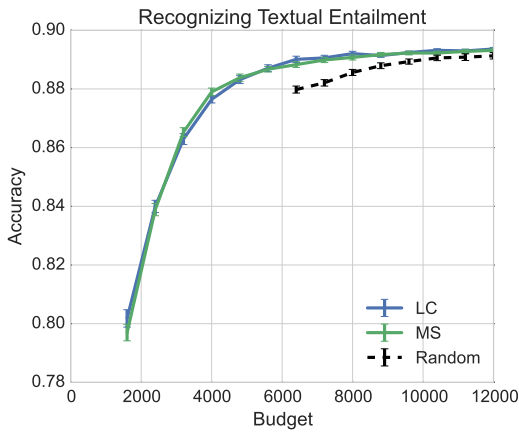}
		\label{fig:RTEComplete}
	}
	\subfigure[$N=204,K=42,S=2$]{
		\includegraphics[width=0.4\textwidth]{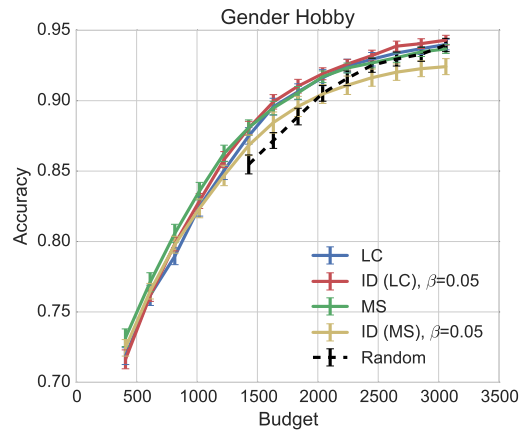}
		\label{fig:GHComplete}
	}
	\caption{Results of different task selection strategies on two real datasets.}
	\label{fig:real-query-comparison}
\end{figure*}

Next, we compare different task selection strategies for BBTA on two real-world datasets.

\subsubsection{Recognizing Textual Entailment}

We first use a real dataset from \emph{recognizing textual entailment} (RTE) tasks in natural language processing. This dataset is collected by using \emph{Amazon Mechanical Turk} (MTurk) \cite{snow08}. For each RTE task in this dataset, the worker is presented with two sentences and given a binary choice of whether the second sentence can be inferred from the first one. The true labels of all tasks are available and used for evaluating the performances of all strategies in our experiments.

In this dataset, there is no context information available, or we can consider all tasks have the same context. That is, this is a homogeneous dataset ($S=1$). The numbers of tasks and workers in this dataset are $N=800$ and $K=164$ respectively. Since the originally collected label set is not complete (i.e. not every worker gives a label for each task), we decided to use a matrix completion algorithm\footnote{We use \emph{GROUSE} \cite{balzano10} for label matrix completion in our experiments.} to fill the incomplete label matrix, to make sure that we can collect a label when any task is assigned to any worker in the experiments. Then we calculate the true accuracy of workers for this dataset, as illustrated in Figure~\ref{fig:rte-workers}. 

Figure~\ref{fig:RTEComplete} depicts the comparison results on the RTE data, showing that LC and MS are comparable, and both of them are better than the random strategy. Since this is a homogeneous dataset, there is only one task type and its proportion is $100\%$. Then ID (LC) and ID (MS) are equivalent to LC and MS respectively and we do not need to plot curves for ID (LC) and ID (MS).

\subsubsection{Gender Hobby Dataset}

The second real dataset we use is \emph{Gender Hobby} (GH) collected from MTurk \cite{mo13}. Tasks in this dataset are binary questions that are explicitly divided into two contexts ($S=2$): sports and makeup-cooking. This is a typical heterogeneous dataset, where there are $N=204$ tasks ($102$ per context) and $K=42$ workers. Since the label matrix in the original GH data is also incomplete, we use the matrix completion algorithm again to fill the missing entries. Figure~\ref{fig:gh-workers} illustrates the distribution of the true accuracy of workers in this dataset. It is easy to see that the labels given by the workers in this dataset are more unreliable than those in the RTE data (Figure~\ref{fig:rte-workers}) due to the increased diversity of tasks.

Figure~\ref{fig:GHComplete} plots the experimental results, showing that ID (LC) performs the best on this typical heterogeneous dataset, and LC follows.

\section{Conclusion}
\label{sec:con}
In this paper, we investigated task selection strategies for task assignment in heterogeneous crowdsourcing. While the existing method BBTA focused on the worker selection strategy in task assignment,  we further extended it by adopting different task selection strategies for picking tasks. The experimental results showed that the performance of BBTA can be further improved by using more appropriate strategies, such as least confidence (LC) and its information density (ID) variant. ID involves tuning the importance parameter $\beta$, which could be cumbersome in practice, while LC does not require tuning any parameter, and its performance is comparable to that of ID in most cases. Therefore, a practical choice is to adopt LC as the task selection strategy in BBTA. 

\section*{Acknowledgment}
HZ was supported by the MEXT scholarship. MS was supported by the CREST program.


\begin{thebibliography}{12}
	
	\bibitem{arora12}
	S.~Arora, E.~Hazan, and S.~Kale.
	\newblock The multiplicative weights update method: a meta-algorithm and
	applications.
	\newblock {\em Theory of Computing}, 8(6):121--164, 2012.
	
	\bibitem{balzano10}
	L.~Balzano, R.~Nowak, and B.~Recht.
	\newblock Online identification and tracking of subspaces from highly
	incomplete information.
	\newblock In {\em Communication, Control, and Computing (Allerton2010)}, pages
	704--711, 2010.
	
	\bibitem{bubeck12}
	S.~Bubeck and N.~Cesa-Bianchi.
	\newblock Regret analysis of stochastic and nonstochastic multi-armed bandit
	problems.
	\newblock {\em Foundations and Trends in Machine Learning}, 5(1):1--122, 2012.
	
	\bibitem{cesa-bianchi06}
	N.~Cesa-Bianchi and G.~Lugosi.
	\newblock {\em Prediction, Learning, and Games}.
	\newblock Cambridge University Press, 2006.
	
	\bibitem{chen13}
	X.~Chen, Q.~Lin, and D.~Zhou.
	\newblock Optimistic knowledge gradient policy for optimal budget allocation in
	crowdsourcing.
	\newblock In {\em Proceedings of the 30th International Conference on Machine
		Learning (ICML2013)}, pages 64--72, 2013.
	
	\bibitem{dawid79}
	A.~P. Dawid and A.~M. Skene.
	\newblock Maximum likelihood estimation of observer error-rates using the em
	algorithm.
	\newblock {\em Journal of the Royal Statistical Society. Series C (Applied
		Statistics)}, 28(1):20--28, 1979.
	
	\bibitem{dempster77}
	A.~P. Dempster, N.~M. Laird, and D.~B. Rubin.
	\newblock Maximum likelihood from incomplete data via the {EM} algorithm.
	\newblock {\em Journal of the Royal Statistical Society, series B},
	39(1):1--38, 1977.
	
	\bibitem{donmez09}
	P.~Donmez, J.~G. Carbonell, and J.~Schneider.
	\newblock Effciently learning the accuracy of labeling sources for selective
	sampling.
	\newblock In {\em Proceedings of the 15th ACM SIGKDD International Conference
		on Knowledge Discovery and Data Mining (KDD2009)}, pages 259--268, 2009.
	
	\bibitem{ertekin14}
	S.~Ertekin, C.~Rudin, and H.~Hirsh.
	\newblock Approximating the crowd.
	\newblock {\em Data Mining and Knowledge Discovery}, 28(5-6):1189--1221, 2014.
	
	\bibitem{finin10}
	T.~Finin, W.~Murnane, A.~Karandikar, N.~Keller, J.~Martineau, and M.~Dredze.
	\newblock Annotating named entities in twitter data with crowdsourcing.
	\newblock In {\em NAACL HLT 2010 Workshop on Creating Speech and Language Data
		with Amazon's Mechanical Turk}, pages 80--88, 2010.
	
	\bibitem{howe08}
	J.~Howe.
	\newblock {\em Crowdsourcing: Why the Power of the Crowd Is Driving the Future
		of Business}.
	\newblock Crown Publishing Group, 2008.
	
	\bibitem{kajino12}
	H.~Kajino, Y.~Tsuboi, and H.~Kashima.
	\newblock A convex formulation for learning from crowds.
	\newblock In {\em Proceedings of the 26th AAAI Conference on Artificial
		Intelligence (AAAI2012)}, pages 73--79, 2012.
	
	\bibitem{karger11}
	D.~Karger, S.~Oh, and D.~Shah.
	\newblock Iterative learning for reliable crowdsourcing systems.
	\newblock In {\em Advances in Neural Information Processing Systems 24
		(NIPS2011)}, pages 1953--1961, 2011.
	
	\bibitem{law11}
	E.~Law and L.~von Ahn.
	\newblock {\em Human Computation}.
	\newblock Morgan \& Claypool Publishers, 2011.
	
	\bibitem{lewis94}
	D.~D. Lewis and W.~A. Gale.
	\newblock A sequential algorithm for training text classifiers.
	\newblock In {\em Proceedings of the 17th Annual International ACM SIGIR
		Conference on Research and Development in Information Retrieval (SIGIR1994)},
	pages 3--12, 1994.
	
	\bibitem{liu13}
	Q.~Liu, A.~Ihler, and M.~Steyvers.
	\newblock Scoring workers in crowdsourcing: How many control questions are
	enough?
	\newblock In {\em Advances in Neural Information Processing Systems 26
		(NIPS2013)}, pages 1914--1922, 2013.
	
	\bibitem{liu12}
	Q.~Liu, J.~Peng, and A.~Ihler.
	\newblock Variational inference for crowdsourcing.
	\newblock In {\em Advances in Neural Information Processing Systems 25
		(NIPS2012)}, pages 701--709, 2012.
	
	\bibitem{mo13}
	K.~Mo, E.~Zhong, and Q.~Yang.
	\newblock Cross-task crowdsourcing.
	\newblock In {\em Proceedings of the 19th ACM SIGKDD International Conference
		on Knowledge Discovery and Data Mining (KDD2013)}, pages 677--685, 2013.
	
	\bibitem{raykar10}
	V.~C. Raykar, S.~Yu, L.~H. Zhao, G.~H. Valadez, C.~Florin, L.~Bogoni, and
	L.~Moy.
	\newblock Learning from crowds.
	\newblock {\em Journal of Machine Learning Research}, 11:1297--1322, April
	2010.
	
	\bibitem{ritter11}
	A.~Ritter, S.~Clark, Mausam, and O.~Etzioni.
	\newblock Named entity recognition in tweets: An experimental study.
	\newblock In {\em Proceedings of the Conference on Empirical Methods in Natural
		Language Processing (EMNLP2011)}, pages 1524--1534, 2011.
	
	\bibitem{scheffer01}
	T.~Scheffer, C.~Decomain, and S.~Wrobel.
	\newblock Active hidden markov models for information extraction.
	\newblock In {\em Proceedings of the International Conference on Advances in
		Intelligent Data Analysis (CAIDA2001)}, pages 309--318, 2001.
	
	\bibitem{settles09}
	B.~Settles.
	\newblock Active learning literature survey.
	\newblock Computer Sciences Technical Report 1648, University of
	Wisconsin--Madison, 2009.
	
	\bibitem{settles08}
	B.~Settles and M.~Craven.
	\newblock An analysis of active learning strategies for sequence labeling
	tasks.
	\newblock In {\em Proceedings of the Conference on Empirical Methods in Natural
		Language Processing (EMNLP2008)}, pages 1070--1079, 2008.
	
	\bibitem{snow08}
	R.~Snow, B.~O. Connor, D.~Jurafsky, and A.~Y. Ng.
	\newblock Cheap and fast - but is it good? evaluating non-expert annotations
	for natural language tasks.
	\newblock In {\em Proceedings of the Conference on Empirical Methods in Natural
		Language Processing (EMNLP2008)}, pages 254--263, 2008.
	
	\bibitem{welinder10b}
	P.~Welinder, S.~Branson, S.~Belongie, and P.~Perona.
	\newblock The multidimensional wisdom of crowds.
	\newblock In {\em Advances in Neural Information Processing Systems 23
		(NIPS2010)}, pages 2424--2432, 2010.
	
	\bibitem{welinder10a}
	P.~Welinder and P.~Perona.
	\newblock Online crowdsourcing: rating annotators and obtaining costeffective
	labels.
	\newblock In {\em Workshop on Advancing Computer Vision with Humans in the Loop
		at CVPR}, pages 25--32, 2010.
	
	\bibitem{zhang15}
	H.~Zhang, Y.~Ma, and M.~Sugiyama.
	\newblock Bandit-based task assignment for heterogeneous crowdsourcing.
	\newblock To appear in \emph{Neural Computation}, 2015. Available at
	http://arxiv.org/abs/1507.05800.
	
	\bibitem{zhou12}
	D.~Zhou, S.~Basu, Y.~Mao, and J.~Platt.
	\newblock Learning from the wisdom of crowds by minimax entropy.
	\newblock In {\em Advances in Neural Information Processing Systems 25
		(NIPS2012)}, pages 2204--2212, 2012.
	
\end{thebibliography}
\end{document}